\documentclass[journal]{IEEEtran}
\usepackage[utf8]{inputenc}
\usepackage{soul}
\usepackage{lettrine}
\usepackage{todonotes}
\usepackage{cite}
\usepackage{supertabular,booktabs}
\usepackage{float}
\usepackage{comment}
\usepackage{makecell}
\usepackage{graphicx}
\usepackage{pdfpages}
\usepackage{amsmath}
\usepackage{algorithm}
\usepackage{algpseudocode}
\usepackage{subfig}
\usepackage{amssymb}
\usepackage{relsize}
\newcolumntype{L}{>{\centering\arraybackslash}m{6.5cm}}
\makeatletter
\newcommand\fs@betterruled{%
  \def\@fs@cfont{\bfseries}\let\@fs@capt\floatc@ruled
  \def\@fs@pre{\vspace*{8pt}\hrule height.8pt depth0pt \kern2pt}%
  \def\@fs@post{\kern2pt\hrule\relax}%
  \def\@fs@mid{\kern2pt\hrule\kern2pt}%
  \let\@fs@iftopcapt\iftrue}
\floatstyle{betterruled}
\restylefloat{algorithm}

\usepackage{hyperref}
\hypersetup{hidelinks}

\def\BibTeX{{\rm B\kern-.05em{\sc i\kern-.025em b}\kern-.08em
    T\kern-.1667em\lower.7ex\hbox{E}\kern-.125emX}}
    
\algrenewcommand\algorithmicrequire{\textbf{Input:}}
\algrenewcommand\algorithmicensure{\textbf{Output:}}
\algnewcommand\algorithmicforeach{\textbf{for each}}
\algdef{S}[FOR]{ForEach}[1]{\algorithmicforeach\ #1\ \algorithmicdo}

\begin{document}

\title{FedMint: Intelligent Bilateral Client Selection in Federated Learning with Newcomer IoT Devices}
\author{
    \IEEEauthorblockN{Osama Wehbi$^1$, Sarhad Arisdakessian$^2$, Omar Abdel Wahab$^2$, Hadi Otrok$^{3}$, Safa Otoum$^{4}$, Azzam Mourad$^{1,5}$ and Mohsen Guizani$^{6}$}\\
    \IEEEauthorblockA{\normalsize$^1$Cyber Security Systems and Applied AI Research Center, Department of CSM, Lebanese American University, Lebanon}\\
    \IEEEauthorblockA{$^2$Department of Computer and Software Engineering, Polytechnique Montréal, Montreal, Quebec, Canada}\\
        \IEEEauthorblockA{$^3$Department of EECS, Center of Cyber-Physical Systems (C2PS), Khalifa University, Abu Dhabi, UAE}\\
    \IEEEauthorblockA{$^4$College of Technological Innovation, Zayed University, Dubai, UAE}\\
    \IEEEauthorblockA{$^5$Division of Science, New York University, Abu Dhabi, UAE}\\
    \IEEEauthorblockA{$^6$Mohammad Bin Zayed University of Artificial Intelligence, Abu Dhabi, UAE}\\
    \IEEEauthorblockA{
    \{\href{mailto:osama.wehbi@lau.edu.lb}{osama.wehbi},
    \href{mailto:azzam.mourad@lau.edu.lb}{azzam.mourad}\}@lau.edu.lb,
    \{\href{mailto:sarhad.arisdakessian@polymtl.ca}{sarhad.arisdakessian},
    \href{mailto:omar.abdul-wahab@polymtl.ca}{omar.abdul-wahab}\}@polymtl.ca,
    \href{mailto:safa.otoum@zu.ac.ae}{safa.otoum}@zu.ac.ae,
    \href{mailto:hadi.otrok@ku.ac.ae}{hadi.otrok}@ku.ac.ae,
    \href{mailto:mohsen.guizani@mbzuai.ac.ae}{mohsen.guizan}@mbzuai.ac.ae
    }
}

\maketitle

\begin{abstract}
Federated Learning (FL) is a novel distributed privacy-preserving learning paradigm, which enables the collaboration among several participants (e.g., Internet of Things devices) for the training of machine learning models.
However, selecting the participants that would contribute to this collaborative training is highly challenging. Adopting a random selection strategy would entail substantial problems due to the heterogeneity in terms of data quality, and computational and communication resources across the participants. Although several approaches have been proposed in the literature to overcome the problem of random selection, most of these approaches follow a unilateral selection strategy. In fact, they base their selection strategy on only the federated server's side, while overlooking the interests of the client devices in the process. To overcome this problem, we present in this paper \textit{FedMint}, an intelligent client selection approach for federated learning on IoT devices using game theory and bootstrapping mechanism. Our solution involves the design of: (1) preference functions for the client IoT devices and federated servers to allow them to rank each other according to several factors such as accuracy and price, (2) intelligent matching algorithms that take into account the preferences of both parties in their design, and (3) bootstrapping technique that capitalizes on the collaboration of multiple federated servers in order to assign initial accuracy value for the newly connected IoT devices. Based on our simulation findings, our strategy surpasses the \textit{VanillaFL} selection approach in terms of maximizing both the revenues of the client devices and accuracy of the global federated learning model.

\end{abstract}
\vspace{5pt}

\section{Introduction}
\label{sec:intro}

\lettrine{T}{h}e adoption and popularity of IoT devices is surging day after day. Based on reports from the International Data Corporation (IDC), by $2025$ the world will contain around 41.6 billion IoT devices. These devices generate large amounts of data. By taking advantage of the heterogeneity and affluence of these data, businesses have the chance to improve their production and business strategies and hence increase their profits. However, most of the times, the common strategy for analyzing IoT data is to gather the data from the devices and to offload these data to a central server for training and pattern extraction \cite{wahab2022intrusion, wang2021toward}. This might not necessarily be scalable in the light of the exponential growth of IoT devices and significant data heterogeneity over the devices \cite{nguyen2021federated, elayan2021sustainability,sorkhoh2022optimizing}. Additionally, considering the vast volume of pervasive IoT data-sets in the big data age \cite{index2016forecast}, coupled with the IoT devices resource constrained nature~\cite{pei2022knowledge, li2021resource}, it is becoming harder to move large amounts of data over the network to cloud data centers for centralized analysis \cite{sorkhoh2020infrastructure, rathee2022trustsys}. Another major concern with such an analysis strategy relates to the privacy risks stemming from the sharing of the data with third-party servers~\cite{rasheed2022explainable, wazzeh2022privacy, otoum2022federated}. This is problematic, especially if there is sensitive information in the training data.

To mitigate these communication and privacy concerns, the Federated Learning (FL) concept has been recently proposed~\cite{rahman2020internet}. The main idea of FL is to perform the training at the level of each device locally, in a distributed fashion~\cite{tabassum2022fedgan, rjoub2022trust}. Google originally unveiled the concept of FL in 2016, when it was applied to Google keyboards to collaboratively learn from numerous Android smartphones \cite{konevcny2016federated} \cite{wahab2021federated}. FL has the potential to revolutionize data analytics in several vital fields, including healthcare, transportation, finance, and smart homes, because it can be applied to any edge device~\cite{qayyum2022collaborative}. A typical Federated training procedure is separated into numerous communication rounds, which are terminated once the global model achieves the intended accuracy \cite{abdulrahman2020survey}. The federated server (i.e, the edge server executing the FL process) first generates a generic machine learning model. Then, The server sends the global model parameters to a set of selected client devices throughout each communication cycle.
The clients train the model on their local data and send the updated parameters to the federated server, which in its turn aggregates the updated parameters to form the new global model \cite{kairouz2021advances}, and returns it to the clients. This procedure is repeated until the intended amount of accuracy is reached or a certain number of communication rounds are achieved.

\subsection{Problem Statement}
\label{problemstatement}

However, in the default federated learning process, the clients that participate in the learning rounds are selected in a random fashion \cite{mcmahan2017communication}. This might be problematic for two main reasons. First, in order for the training to take place and for the results to be communicated with the federated server, the IoT devices (hereafter, interchangeably used with \textit{clients}) need to dedicate appropriate amounts of resources such as CPU, RAM, and bandwidth to be able to both train the model and transmit it efficiently~\cite{albaseer2022balanced, hammoud2022demand}. Nonetheless, this criterion is often not realized due to the resource limits of certain IoT devices, which have low computing and communication capabilities, causing considerable latencies to the synchronized parameter aggregation process at the server \cite{nguyen2021federated}. Second, owing to the heterogeneity at the level of the clients in federated learning, different clients might have different types of datasets with varying sizes, qualities, and distributions, a problem that is often referred to as the non-IID (non-independent and identically distributed) problem. Thus, selecting clients at random can result in having clients with low resources or clients that hold smaller amounts of data~\cite{hammoud2022data}. This might hinder the objective of achieving a certain desired accuracy level and could result in a considerably high number of communication rounds. Furthermore, the problem of newcomer client devices is challenging and has not been appropriately addressed in the literature. The existing of newcomer IoT devices in the environment makes the selection process more complicated. Dealing with devices that can not afford any recorded data for previous interaction that can tell anything about the status of such device in order to build a decision on is very challenging. Moreover, most of the existing client selection models have a unilateral selection mechanism in which the federated servers take the selection decision based on specific standards, which results in an unfair or win-lose situation wherein the server's needs are met while the clients' opinions are completely neglected.

\subsection{Contributions}
\label{contributions}
Motivated by these shortcomings of the random client selection approach, we propose a novel client selection approach for federated learning on IoT devices called \textbf{Fed}erated \textbf{Int}elligent \textbf{M}atching (\textit{FedMint}) which is based on matching game theory \cite{chiti2018matching, roth1992two, arisdakessian2022survey, chiti2020matching} and trust bootstrapping. First, we apply a bootstrapping method to obtain an initial accuracy value for the newly deployed IoT devices. The proposed trust bootstrapping method capitalizes on the collaboration between an active set of federated servers and a central bootstrapping server, wherein each active federated  server contributes its recorded, that records the server’s interactions with a group of client IoT devices in the previous training rounds. In return, the bootstrapping server trains a Decision Tree model on the collected data and finally, delivers the predicted accuracy of the newcomer IoT device to the federated server that send the inquiry. Moreover, we employ a matching game wherein clients and federated servers with specific FL tasks are given the chance to form their own preference lists based on certain criteria. Using these lists, the matching takes place where at each FL communication round. Unlike the random selection approach, our method makes the federated servers aware of the data types that are offered by the IoT devices as well as their accuracy levels. The preferences of the IoT devices are also taken into consideration in the matching process in terms of monetary rewards.

In this work, the main contributions are summarized as
follows:
\begin{itemize}

\item Designing an accuracy bootstrapping model that helps federated servers assign initial accuracy values for the newcomer IoT devices having no past participation's. The proposed bootstrapping model ensures fairness between already active and the newcomer IoT devices in terms of the chance to participate in future FL training rounds.

\item Devising a rewarding model for federated servers to encourage them to engage in the bootstrapping process. This is achieved by setting a limit on the bootstrapping requests that a federated server can make and linking the increase of this limit to the number and quality of contributions made by each server to the bootstrapping phase.

\item Proposing a bilateral client selection approach for federated learning using matching game theory. To the best of our knowledge, this is the first client selection approach in FL that takes into account the preferences of both the federated servers and client devices in the selection process while considering the newcomer IoT devices problem.

\item Elaborating a distinct optimization problem for the federated servers as well as for the client IoT devices while expressing the objectives and constraints of both parties in the selection process.

\item Implementing a set of distributed matching algorithms which takes into account the preferences of both the federated servers and client devices. The proposed algorithms lead to a stable matching point from which neither the servers nor clients have incentive to deviate.

\end{itemize}

\subsection{Paper Outline}
\label{paperoutline}
In section~\ref{sec:Rel}, we study the literature on client selection and discuss the originality of our solution. In section~\ref{sec:problemformulation}, we create the optimization problems for the federated servers and client IoT devices. In section~\ref{sec:boots}, we introduce the bootstrapping mechanism and architecture in addition to the bootstrapping motivation function for the federated servers. In section~\ref{sec:proposed}, we describe Fed-IoT matching game principles and terminology, interpret preference functions, and suggest algorithms for creating preference lists. In section~\ref{matchingselection}, we deliver the distributed version of the matching algorithms implementation. In section~\ref{sec:experiments}, we demonstrate the simulation setup we utilized to run our tests and interpret the outcomes. Finally, we conclude our work in section~\ref{Conclusion}

\section{Related Work}
\label{sec:Rel}
In this section, we review relevant literature on client selection and trust bootstrapping in federated learning.

\subsection{Client Selection}

There are several works in federated learning that focus on client selection.
In \cite{goetz2019active} the authors suggest a probability conditioned client selection mechanism, based on clients' data. In this approach, clients are selected based on a probability calculated using a value returned by an evaluation function on the client's device. The authors argue that their approach minimizes overall training rounds required to attain the desired accuracy level.

In \cite{cho2020client}, the authors introduce POWER-OF-CHOICE, a client selection framework that achieves a balance between solution bias and convergence speed in a flexible manner. The outcomes reveal that this approach converges $3x$ faster and achieves better test accuracy compared to the normal random selection by almost $10\%$ .

In \cite{nishio2019client}, the authors address the challenge of limited computational resources on client devices and propose an approach called FedCS. The proposed approach seeks to mitigate the problem of resource variety across client devices through performing the federated learning tasks efficiently, by properly managing clients in accordance with their resources conditions. This is done through imposing some constraints on the updated models that should be accepted. 

In \cite{abdulrahman2020fedmccs} The authors introduce FedMCCS, a multi-criteria methodology targeting the client selection process in FL to address the heterogeneity of client devices. This approach takes into account a device's resources. The resources are assessed to predict whether or not the device can accommodate a FL task. 

In \cite{huang2020efficiency}, the authors propose an approach, called RBCS-F, which advocates a fairness-guaranteed algorithm. The algorithm seeks to establish a suitable balance among both training efficiency and fairness, while minimizing the average model exchange time. This approach ensures that trainers with low importance are not neglected from participating in the federated training process. 

In \cite{zhao2022participant}, the authors introduce \textit{Newt}, a novel client selection approach that investigates a trade-off between accuracy and system advancement. As a novel feature of client selection technique design, a control on selection frequency is incorporated in the approach.

The above-discussed papers take into consideration the federated server's only, while disregarding the needs and preferences of the clients. This would result in biased and sometimes unfair scenarios, where the federated server has the ultimate say as to which clients should be selected. On the contrary, our strategy considers the preferences and restrictions of both the federated server and client devices in the selection process, to guarantee more fair and less biased decisions.

In \cite{chen2021matching}, a matching-theoretic method in multi-access edge computing network with incomplete preference list was developed to handle the low-latency task scheduling problem. The matching occurs between the edge nodes in charge of the federated learning task and the end devices in a wide environment. Experiments reveal that the complete preference list matching approach performs slightly better than the matching approach by reducing the latency due to the missing information.
To manage federated learning task allocation and defend against malicious clients, \cite{kang2020training} applied a modified one-to-one two-sided matching theory between workers and task publishers, as well as a worker reputation metric. The proposed solution focuses on minimizing the job publisher's training time as well as the worker's energy usage. In many aspects, our strategy differs from the above mentioned two solutions. Both techniques attempt to minimize particular metrics, but our model maximizes client incentives as well as federated server global model accuracy. We are also applying a distributed version of the matching game theory, which is more compatible with the distributed nature of the federated learning. Furthermore, none of the articles cited above addressed the issue of newcomer IoT devices in FL.


\subsection{Trust Bootstrapping}
Trust bootstrapping have been used to solve the recommender system cold-start problem in many fields such as business, networking, cloud services and others. In \cite{cao2020fltrust} the authors targeting the malicious client aspect in FL by introducing FLTrust. Unlike the normal byzantine-robust methods that rely on statistics analysis for malicious clients detection. The proposed approach counts on a small amount of collected data to bootstrap trust. 

In \cite{dong2021flod}, the authors introduce FLOD. As a new byzantine-resistant federated learning approach. The proposed method capitalizes on trust bootstrapping and the hamming distance based aggregation, in addition to additive homomorphic encryption and multiple optimizations to protect privacy and byzantine-robustness in federated learning. 

The above mentioned approaches named as FLTrust and FLOD, proved that using trust bootstrapping to assign initial trust score for each local model is effective compared to other methods. These solutions, unlike ours, are aimed at the bootstrapping of the federated model rather than newcomer IoT devices. 

In  \cite{wahab2022federated}, the authors try to solve the recommendation system cold start problem using federated learning by introducing a double deep Q learning model that counts on the IoT devices trust score, as well as the resources availability in the selection process. In this approach, the author applied the federated learning to address the bootstrapping while in this paper we are using bootstrapping to address the problem of newcomer IoT devices.

\section{Problem Formulation}
\label{sec:problemformulation}
In this section, we express the client IoT devices selection problem in FL as an optimization problem and clarify the relevant constraints. Note that we describe the various symbols utilized throughout this work in TABLE \ref{tab:TABLE I}.

\begin{table}[!ht]
\caption{LIST OF DEFINITIONS}
\begin{center}
\begin{tabular}{|c L|}
\hline 
 & \\
Symbol & Explanation \\ [0.5ex] 
\hline\hline
 & \\
 &  \textbf{IoTs and Federated Servers}\\
$s$ & A single federated server\\
$i$ & A single IoT device\\
$\varphi_s$ & Data type requested by $s$\\
$\varphi_i$ &  Available data on $i$\\
$p_w$ & Resource $w$'s unit price offered by $s$\\
$S$ & Set of active federated servers\\
$I$ & Set of active IoT devices\\
$L_{i s}$ & Scaled network latency between $i$ and $s$\\ 
$r_{n}$ & A federated server communication round $n$\\
$C_s$ & Number of clients requested by a federated server\\
$N_s$ & Number of clients selected by a federated server\\
$A_{N_s}$ & List of clients selected by a federated server $s$ \\
$CPU_{pro_i}$ & Amount of CPU resources promised by an IoT device $i$ \\
$RAM{pro_i}$ & Amount of RAM resources promised by an IoT device $i$  \\
$Band{pro_i}$ & Amount of bandwidth resources promised by an IoT device $i$ \\
$e{_i}(o)$ & Operational earnings of an IoT device $i$\\
$e{_i}(t)$ & Communications and traffic earnings of an IoT device $i$\\
$\beta^{i}_{s}$& A request from federated server $s$ to the bootstrapping server to inquire about a newcomer IoT device $i$ \\
& \\
& \textbf{Optimization}\\
$E(i)$ & Total Monetary reward of an IoT device $i$\\
$Acc{_i}$ & Local testing accuracy of an IoT device $i$\\
$Acc{_s}$ & Global Model Accuracy of a federated server $s$\\
$\widehat{Acc_i}$ & Predicted local testing accuracy of an IoT device $i$\\
& \\
& \textbf{Matching Theory}\\
$\gamma$ & A Matching relation between two entities\\ 
$\gamma(s)$ & A matching scheme of a federated server\\
$\gamma(i)$ & A matching scheme of an IoT device\\
$P_i$ & IoT device $i$ preference list\\
$P_s$ & The preference list of a federated server $s$\\
$i_1 \succ_s i_2$ & Server $s$ prefers being matched to $i_1$ rather than $i_2$\\
& \\
& \textbf{Bootstrapping}\\
$SD(X)$ & Standard Deviation of an attribute X\\
$SD(Y, X)$ & Standard Deviation of two attributes X and Y\\
$CV$ & Coefficient of Variation\\
$P(c)$ & Probability of a category c\\
$\eth$ & Central Bootstrapping Server\\
$n$ & Sample Size\\
$\overline{x}$ & Average of a sample\\
$SDR(Y, X)$ & Standard Deviation Reduction\\
$Calls(s)$ & Bootstrapping calls that s can make\\
&\\
\hline
\end{tabular}
\end{center}
\label{tab:TABLE I}
\end{table}

\subsection{Client IoT Device Optimization Problem}
\label{sec:IoTOpt}

The primary goal of IoT devices is to increase their profitability. The earnings $E(i)$ of every IoT device are determined based on the amount of resources that the device promises to dedicate for the FL process (i.e, $cpu_{pro_i}$, $ram_{pro_i}$, and $band_{pro_i}$). The earnings of an IoT device is the summation of two functions, namely (1) operational and (2) network traffic earnings.

\begin{enumerate}

    \item \textbf{Operational Earnings}: The operational earnings of an IoT device are made up of two measures, i.e., $CPU{pro_i}$ and $RAM{pro_i}$. The CPU and RAM utilization cost (in MIPS) measures the amount of CPU and RAM used by a specific IoT device $i$ when performing federated server operations. Formally, the operational earnings $e{_i}(o)$ of $i \in I$ are defined as stated below:
        \begin{equation}\label{eq:1}
            e{_i}(O) = CPU{pro_i}*p_w + RAM{pro_i}*p_{w^\prime}
        \end{equation}

    \item \textbf{Traffic Earnings:} In FL, the devices must send/receive the model parameters to/from the federated servers at certain bandwidth rates. Depending on the underlying demand, the active physical links may be unavailable at different periods. Thus, the traffic earnings of IoT devices is estimated as the bandwidth earnings incurred on the $s \Leftrightarrow i$ link multiplied by the link's scaled undergoing delay as a penalty.
    
    In formal terms, the traffic earnings $e{_i}(t)$ of an IoT device $i \in I$ interacting with a federated server $s \in S$ is defined as follows:
    
        \begin{equation}\label{eq:2}
            e{_i}(t) = (Band{pro_i} * p_w) \times (1 - L_{i s})
        \end{equation}
\end{enumerate}

As a result, each IoT device $i \in I$ must maximize the objective function in Eq~(\ref{eq:3}), where \textit{std} represents the standard deviation of the IoT device $i$'s local accuracy compared to the federated server's global model accuracy of $s \in S$ at round $r_n$. 
Such a multiplication reflects the fact that each IoT device will be penalized in terms of the gap between its local accuracy and the overall global model accuracy. Therefore, as the IoT device's local accuracy $Acc_i$ tends to be closer to the global model's accuracy $Acc_s$, the IoT device will receive a higher reward.

        \begin{equation}
        \label{eq:3}
        E(i) = (e{_i}(O) +e{_i}(t)) \times (1 - std_{r_n})
        \end{equation}
        

\textbf{Constraint 1}: Each IoT device $i \in I$ can be matched with only one single federated server, per each communication round.
        \begin{equation}
        \label{eq:4}
        0 \leq |\gamma(i)| \leq 1
        \end{equation}

\subsection{Federated Servers Optimization Problem}
\label{sec:FedOpt}

Federated servers are interested in maximizing the accuracy $Acc_s$ of the deep learning model, by selecting the most appropriate set of client devices in terms of historical accuracy. 
The accuracy $Acc_s$ of the FL training process can be derived as per \eqref{eq:5}: 
        \begin{equation} \label{eq:5}
            Acc_s = \frac{\sum_{n = i}^{I^\prime} weighted\_accuracy_{n}}{\sum_{n = i}^{I^\prime} test\_data\_size_{n}}
        \end{equation}
        
where $I^\prime$ represents subset of IoT devices of $I^\prime \subseteq I$ that participated in federated learning round $r_n$ with federated server $s$. Weighted\_accuracy can be calculated for each IoT i $\in$ I as follows :
        \begin{equation} \label{eq:6}
            weighted\_accuracy_i = Acc_i * test\_data\_size_i
        \end{equation}
In this way, a federated server $s$ will be more interested in selecting an IoT device $i$ over device $i^\prime$ in a certain communication round if and only if:
        \begin{equation} \label{eq:7}
            Acc_i \succ_s Acc_{i^\prime} 
        \end{equation}
\textbf{Constraint 2}: The selected client IoT devices total number should not exceed the amount of requested IoT $C_s$ by $s$.
        \begin{equation} \label{eq:8}
            N_s \leq C_s 
        \end{equation}

\section{Bootstrapping Accuracy for Newcomer IoT Devices}
\label{sec:boots}
In this section, we explain the trust bootstrapping solution and discuss the main stages of the process.

Bootstrapping is a recommendation technique used in many fields, such as cloud computing where it is used in order to determine the trust level of the newly deployed cloud services, when no historical records on their previous actions exists~\cite{wahab2020endorsement, wahab2016towards}. In our approach, we are going to apply the bootstrapping in order to assign initial accuracy value for the newcomer IoT devices. In the proposed matching solution, federated servers sort their preference lists of IoT devices depending on the accuracy level of each IoT device. So, devices with no accuracy can't be added to the preference list. Moreover, assigning random accuracies for such devices may result in an unjust situation for both IoT devices and federated servers. This happens when a very good IoT device gets a very low accuracy, and vice versa. The proposed method capitalizes on the collaboration of multiple active federated servers with a central bootstrapping server to help these servers to overcome the problem by providing initial accuracy for newcomer IoT devices.

\subsection{Bootstrapping Overview}
\label{sec:Bootstrapping Overview}
\begin{figure}[h!]
    \centering
    \includegraphics[width=\linewidth]{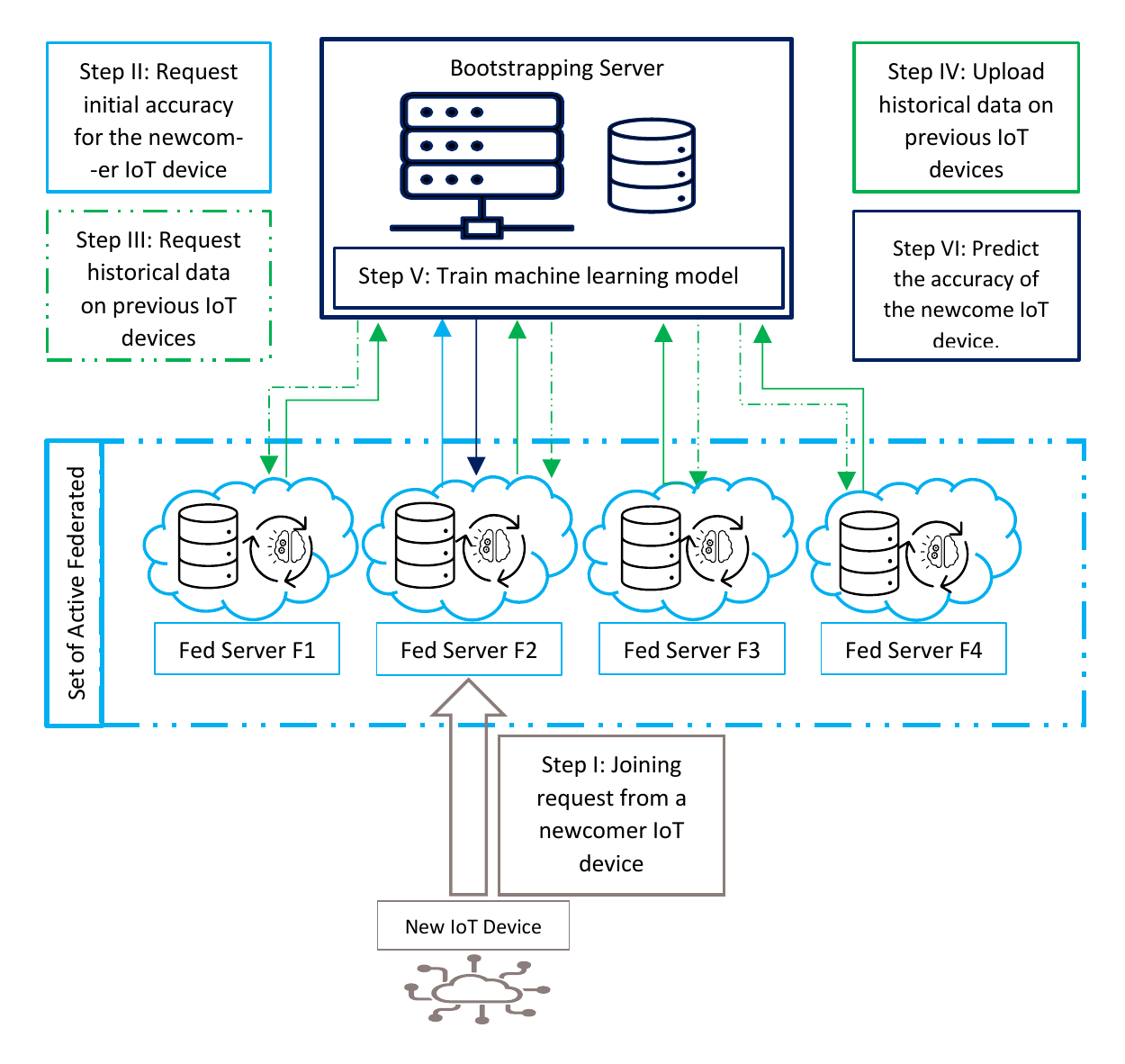}
    \caption{Bootstrapping Architecture}
    \label{fig:boot}
\end{figure}

Our solution is illustrated in Fig.~\ref{fig:boot} and described as follows. Each federated server maintains a dataset that records the server's interactions with a group of client IoT devices in the previous training rounds. This dataset should include the characteristics of each IoT device that participate in the round (i.e., region, type of the device, resources usage) and most importantly the accuracy that this device has achieved in the designated round. In each round, following a joining request from a certain newcomer IoT device (Step I) to an active federated server, the entitled server is going to ask the central bootstrapping server to predict the expected accuracy for the newcomer IoT device (Step II). The bootstrapping server issues a bootstrapping request and broadcasts it to all active federated servers (Step III). Next, the active federated servers are going to upload all/part of their historical dataset to a bootstrapping central server (Step IV). Then, the central server trains a decision tree regression model on the collected datasets to predict the expected accuracy of the newcomer IoT device (Step V). Finally, the bootstrapping server sends the initial expected accuracy of the newcomer IoT device to the asking server (Step VI). Decision tree uses a tree structure to develop regression or classification models. It recursively divides a dataset sample into relatively small segments while also creating a related decision tree. As final outcome, a tree should be created with leaf nodes representing each a numerical target decision, and decision nodes that include at least two branches representing the examined attribute values. The topmost decision node in a tree known as root node, that represents the most reliable predictor that can be chosen based on certain statistical metric (e.g., Standard Deviation Reduction).

\subsection{Decision Tree Creation}

The decision tree model is a well known supervised machine learning model that capitalizes on the ID3 technique for creating decision trees. This is done by performing a greedy top-down search across the range of possible paths without backtracking\cite{quinlan1986induction}. Decision tree uses a tree structure to develop regression or classification models. The major distinction between regression and classification decision trees is that the results of classification-based decision trees are categorical, whereas the results of regression-based decision trees are continuous. The regression decision trees accept both ordered and continuous data. 
The ID3 approach can be applied to create a regression decision tree by using Standard Deviation Reduction (SDR) instead of Information Gain~\cite{quinlan1992learning}. Standard deviation reduction technique is going to be used in order to measure the homogeneity in a feature. 
Decision Tree is created in a top-down way out-of a root node by splitting the data into segments containing homogeneous samples with comparable values. To determine the uniformity of a numerical data sample, we utilize Standard Deviation (SD). A numerical sample with a low Standard Deviation is more likely to be homogeneous, whereas a sample with a high Standard Deviation is less likely to be homogeneous. The SDR is a measure of the decrease in the Standard Deviation after splitting a dataset based on a certain attribute. The most homogeneous branch is determined by the attributes that result in the greatest standard deviation decrease. The SDR of a specific feature X can be derived by subtracting the standard deviation of the target Y, i.e., SD(Y), before the split, from the standard deviation after the split by X SD(Y, X) represented by Eq.(\ref{eq:100})

        \begin{equation}\label{eq:100}
            SDR(Y, X) = SD(Y) - SD(Y, X)
        \end{equation}
        
where a certain SD can be calculated for one attribute $x$ as follows:

        \begin{equation}\label{eq:101}
        SD(x) = \sqrt{\frac{\sum (x-\overline{x})^{2}}{n}}
        \end{equation}
        
The symbol $n$ represents the sample size and $\overline{x}$ represents the sample's mean that can be derived as below:

        \begin{equation}\label{eq:102}
        \overline{x} = \frac{\sum x}{n}
        \end{equation}

The Standard Deviation for two attributes (Target and Predictor) is defined as below:

        \begin{equation}\label{eq:103}
        SD(Y, X) = \sum_{c \in X} P(c)SD(c)
        \end{equation}
        
Then, the Coefficient of Variance (CV) can be computed as per Equation (5):

    \begin{equation}\label{eq:104}
            CV = \frac{SD}{\overline{x}} * 100\%
            \end{equation}
            
\bgroup
\def\arraystretch{1.5}%
\begin{table}[h!]
\caption{Data Sample}
\begin{center}
\begin{tabular}{||c|c|c|c||}
\hline
& & & \\
\textbf{Provider} & \textbf{Region} & \textbf{DeviceType} & \textbf{Accuracy} \\ [0.5ex]\hline
P4       & Asia    & Watch      & 73.69 \\ \hline
P1       & Asia    & Phone      & 65.05 \\ \hline
P4       & America & Security   & 67.62 \\ \hline
P3       & America & Lock       & 58.54 \\ \hline
P1       & America & Phone      & 53.85 \\ \hline
P2       & Africa  & Lock       & 56.37 \\ \hline
P1       & Europe  & Watch      & 53.85 \\ \hline
P4       & America & Security   & 82.42 \\ \hline
P3       & Asia    & Phone      & 95.92 \\ \hline
P1       & Europe  & Watch      & 55.56 \\ \hline
P1       & America & Security   & 56.80 \\ \hline
P2       & Africa  & Watch      & 52.88 \\ \hline
P4       & Asia    & Watch      & 90          \\ \hline
P3       & Asia    & Security   & 55          \\ \hline
\end{tabular}
\end{center}
\label{tab:TABLE II}
\end{table}
\egroup

To demonstrate how Standard Deviation Reduction may be used in practical federated learning scenarios to generate decision trees, we provide an illustrative example using a portion of the dataset that shown in Table~\ref{tab:TABLE II}. This dataset was generated in order to be used subsequently in the experimental analysis. The dataset stores information on the IoT devices that contribute in the federated learning rounds in terms of their type, deployment region, provider and observed accuracy. First of all, we should calculate the SD for the target which means the Accuracy denoted by SD(Accuracy). By applying the above-explained equations on the Accuracy feature we can find that:

\begin{itemize}
    \item $n = 14$
    \item $\overline{x} = \frac{917.55}{14} = 65.53$
    \item $SD(Accuracy) = 13.96$
    \item $CV = 21.31\%$
\end{itemize}

The previous calculations are needed in order to evaluate the splitting impact of each feature. Starting with the Provider feature as illustrated in Table~\ref{tab:TABLE III}, the dataset is grouped by category where the standard deviation for each group is calculated alongside with the frequency of each category.

\bgroup
\def\arraystretch{1.5}%
\begin{table}[h!]
\caption{Split according to the "Provider" feature}
\begin{center}
\begin{tabular}{|cc|c|c|}
\hline
                               &    &                 &       \\
                               &    & Accuracy(SD) & Frequency \\[0.5ex] \hline
\multicolumn{1}{|c|}{Provider} & P1 & 4.16     & 5     \\ \cline{2-4} 
\multicolumn{1}{|c|}{}         & P2 & 1.74     & 2     \\ \cline{2-4} 
\multicolumn{1}{|c|}{}         & P3 & 18.51     & 3     \\ \cline{2-4} 
\multicolumn{1}{|c|}{}         & P4 & 8.50    & 4     \\ \hline
                               &    & SD(Accuracy, Provider) = 8.13                & Total = 14    \\ \hline
\end{tabular}
\end{center}
\label{tab:TABLE III}
\end{table}
\egroup

The Provider feature SDR can be determined by applying the following illustrated methodology:

\begin{itemize}
    \item $SD(Accuracy, Provider)=P(P1)*SD(P1) +\\ P(P2)*SD(P2) + P(P3)*SD(P3) + P(P4)*\\SD(P4)= (5/14)*(4.16) + (2/14)*(1.74) + (3/14)*(18.51) + (4/14)*(8.50) = 8.13$
    \item $SDR(Accuracy, Provider) = SD(Accuracy) -\\SD(Accuracy, Provider) = 13.96 - 8.13 = 5.83$
\end{itemize}

By applying the same concept for all the features (i.e. Region, DeviceType) we will obtain the following standard deviation values:

\begin{itemize}
    \item $SD(Accuracy, Provider)= 5.83$
    \item $SD(Accuracy, Region)= 13.96 - 9.51 = 4.45$
    \item $SD(Accuracy, DevType)= 13.96 - 12.28 = 1.67$
\end{itemize}

\begin{figure}[h!]
    \centering
    \includegraphics[width=\linewidth]{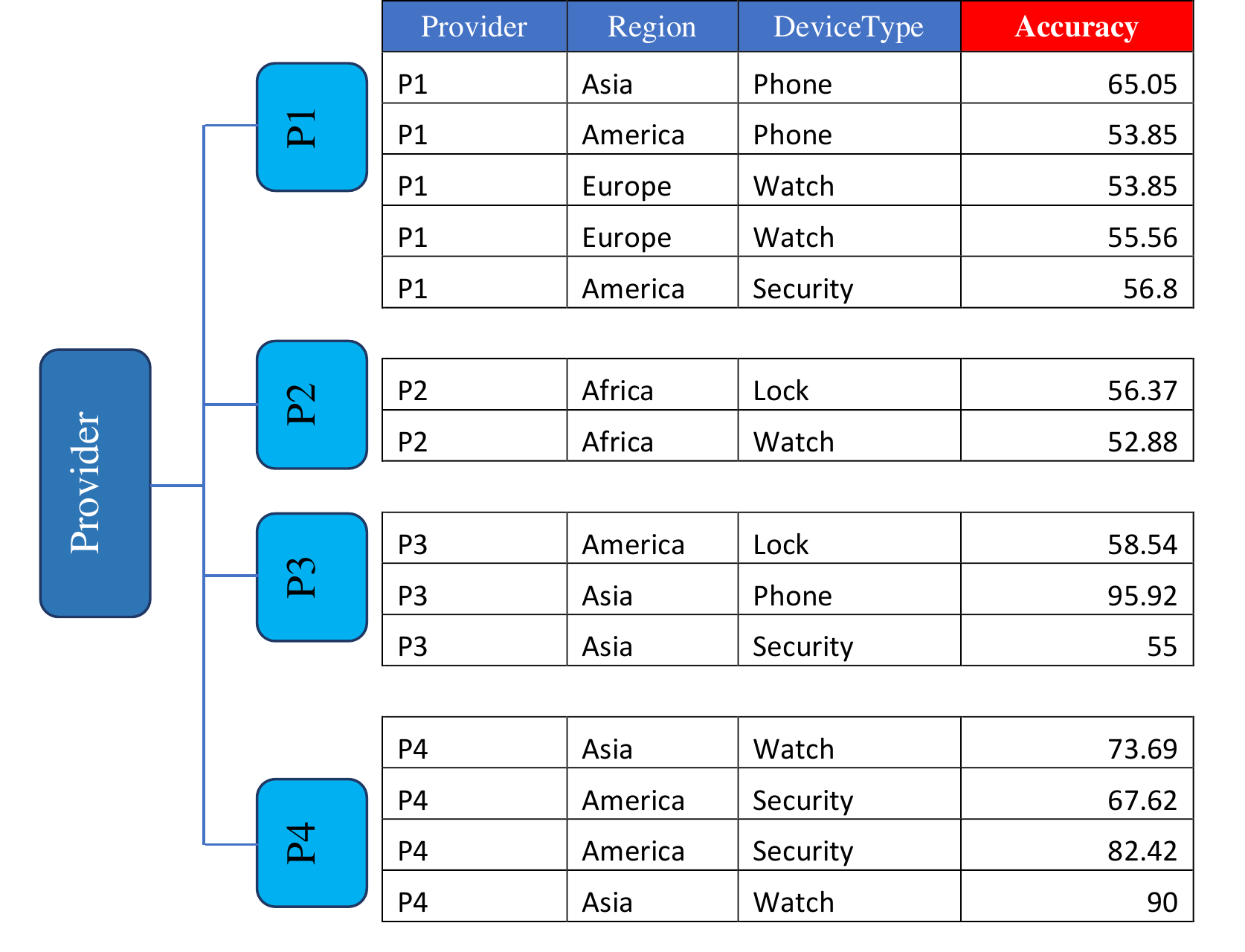}
    \caption{Dataset Splitting Based on Provider Feature}
    \label{fig:firstpar}
\end{figure}

\begin{figure}[h!]
    \centering
    \includegraphics[width=\linewidth]{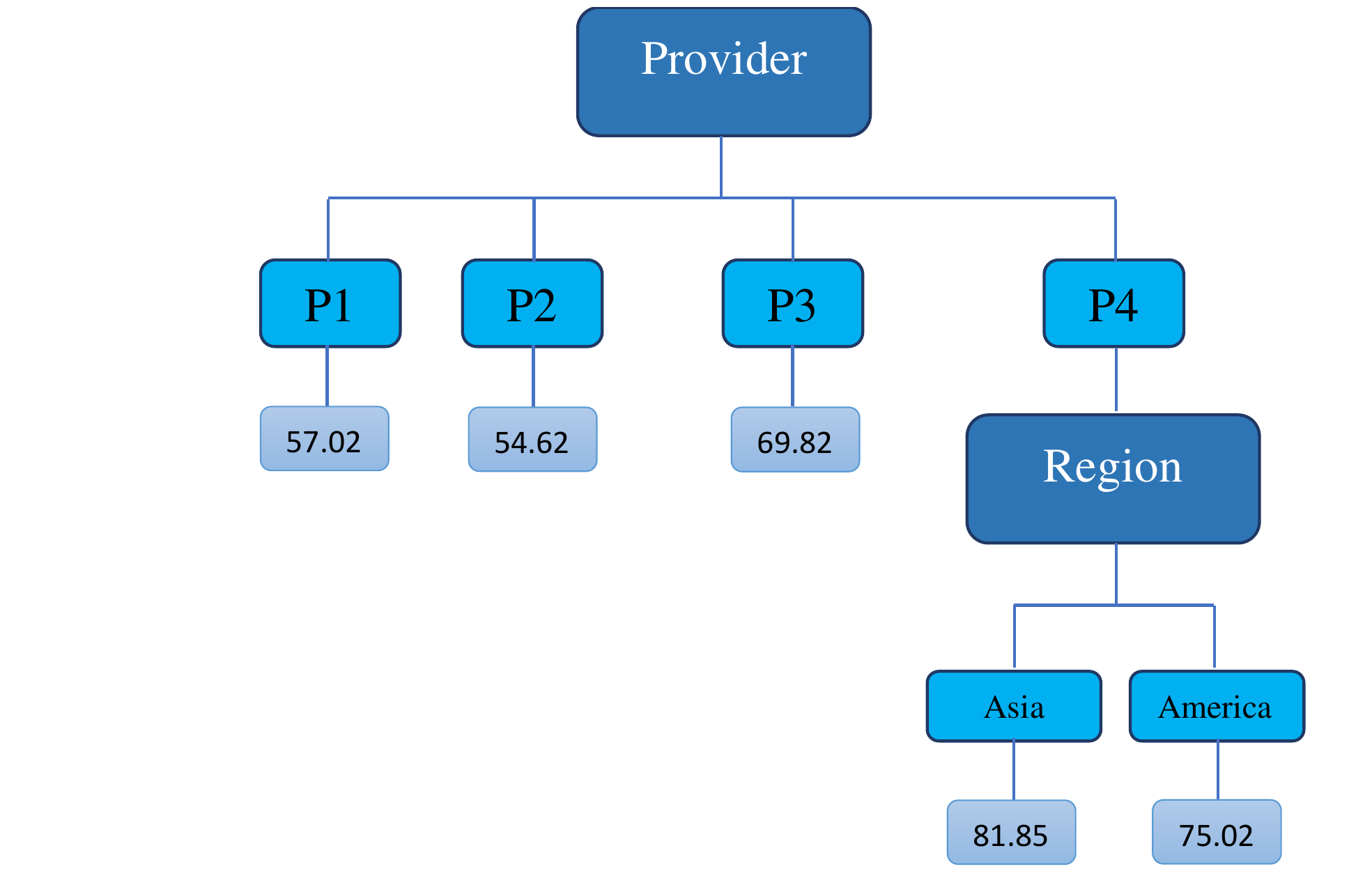}
    \caption{Final Decision Tree (n=3 and CV=10\%)}
    \label{fig:finalpar}
\end{figure}

Based on the calculated values, the root node should be assigned to the attribute with the greatest Standard Deviation Reduction which is the Provider in our example. The dataset is partitioned depending on the values of the Provider feature as shown in Fig.~\ref{fig:firstpar}. This process is repeated on the non-leaf branches until all data has been processed, or until a branch's Coefficient of Variation (CV) falls below a specific threshold and/or few or no more instances remain in the branch. Finally, if there are more than one occurrence at a leaf node, we use the average as the final value for the target. By applying the above rules on our data sample with $n = 3$ as number of instances threshold and $CV = 10\%$ as coefficient of deviation threshold, we obtain the final tree structure illustrated in Fig.~\ref{fig:finalpar}.

\subsection{Motivation function}
Federated servers should be encouraged to engage in the bootstrapping process. Therefore, it is very important to provide a rewarding technique for these servers that have participated in the bootstrapping process as encouragement for them to participate again. We propose Eq.~(\ref{eq:105}), that represents a positive relation between the number of bootstrapping calls that a federated server can make $calls(s^\prime)$, with respect to its cumulative previous number of contributions $Ccont$ and the data rate $DR_t$ that the federated server provides based on the total data set size uploaded at time $t$ by all the participated servers.

        \begin{equation}\label{eq:105}
        \begin{aligned}
        Calls (s^\prime)_t = Calls (s^\prime)_{t-1} + (|Ccont| + |Ccont*DR_t|+1)
        \end{aligned}
        \end{equation}

where $DR_t$ of a specific federated server $s^\prime$ can be calculated as follows:
            \begin{equation} \label{eq:106}
            DR_t = \frac{uploaded\_data\_size_{s^\prime}}{\sum_{n = s}^{S} uploaded\_data\_size_n}
        \end{equation}
        
Establishing win-to-win or rewarding methodology between federated servers and the bootstrapping server is important to guarantee the constant update of the provided data by the federated servers used in the bootstrapping model training. Thus, each federated server is going to be forced to contribute by its data in order to be able to make further bootstrapping requests from the bootstrapping server, to assign accuracy values for new IoT devices that can be potential clients to a certain federated learning round. Such a relationship is beneficial to both parties. In fact the federated servers can benefit by inquiring about newcomer IoT devices and get extra number of eligible bootstrapping calls. On the other hand, the bootstrapping server can benefit from the data provided by the federated servers to keep the Decision Tree model updated.
        
\section{Proposed Approach: Intelligent Client Selection Mechanism}
\label{sec:proposed}
In this section, we describe the proposed architecture, explain the preference functions for both clients and federated servers, and provide the intelligent client selection algorithms.

\subsection{Proposed Architecture and Solution Overview}
\label{}
\begin{figure}[ht]
    \centering
    \includegraphics[width=\linewidth]{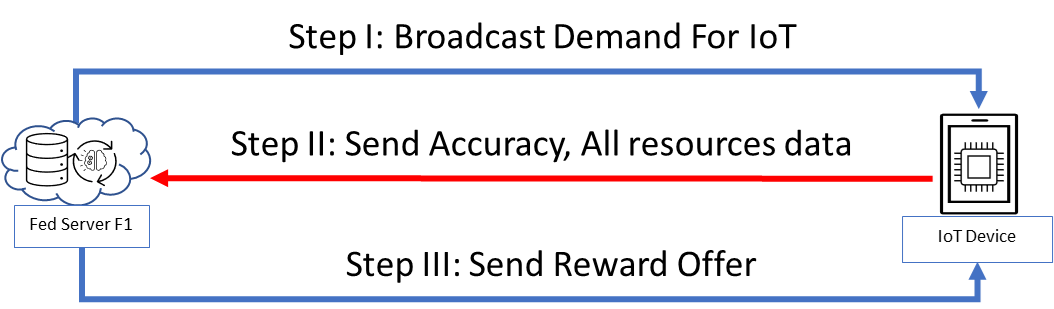}
    \caption{Fed-IoT First Communication}
    \label{fig:CArchi}
\end{figure}

The initial communication round between federated server and client IoT devices is illustrated in Fig.~\ref{fig:CArchi}. This round is important for the distributed intelligent selection approach, as it allows both parties to exchange all the needed information for the selection process. The communication start with a demand request to all the active IoT broadcasted by a specific federated server (Step I), after that the interested/available IoT devices reply to the federated server with a message that contains all the needed information including the IoT device accuracy from previous work (Step II), if the IoT device is newly deployed device then the federated server is going to use the bootstrapping methodology discussed in Section~\ref{sec:boots}. Finally the federated server sends his offer to the designated IoT device (Step III).

Our solution takes two inputs, i.e., a set of federated servers that need to select a set of clients to execute an FL task, and a set of active IoT devices that are ready and willing to participate in the FL process. In the following, we highlight the main steps of our solution:

\begin{enumerate}
    
    \item \textbf{Preference Lists creation}: In this step, each of the clients and federated servers build their preferences lists:
    \begin{itemize}
        \item Initially each active federated server broadcasts to the IoT devices in the environment the requirements of each FL task. The IoT devices that are active and willing to participate reply by sending to the server their accuracy values obtained from their participation in previous FL tasks or from the bootstrapping server alongside their resource information.
        
        \item Client IoT device preference list: Contains the list of federated servers sorted based on the reward values that these servers offer to pay for the IoT device. IoT devices prioritize the federated servers that offer higher rewards.
        
        \item  Federated server preference list: Contains the list of IoT devices sorted according to the accuracy values of the IoT devices. Federated servers prioritize the clients that have higher accuracy values.
    \end{itemize}
   \begin{figure}[ht]
    \centering
    \includegraphics[width=\linewidth]{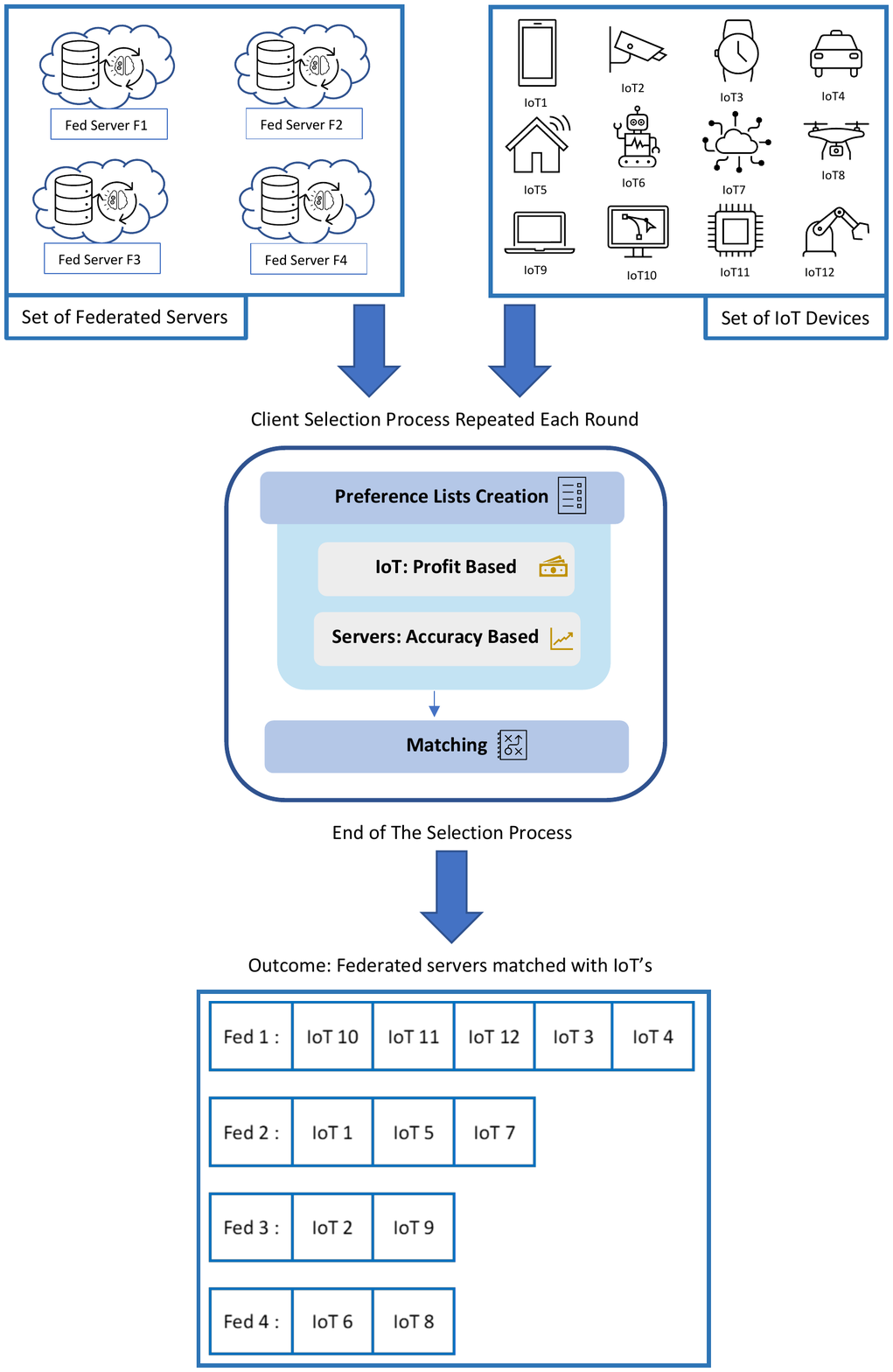}
    \caption{Architecture of the proposed approach}
    \label{fig:Archi}
    \end{figure} 
    \item \textbf{Matching}: In this step, the matching between the federated servers and client IoT devices takes place. The matching is accomplished based the matching algorithms which we describe in section \ref{matchingselection}. The algorithms rely on the preference lists of the IoT devices and federated servers which we describe in Sections \ref{IoTMatching} and \ref{ServerMatching} respectively. The aim of this step is to reach a stable matching point wherein each IoT device is matched to a federated server and both parties do not have any incentive to deviate from this matching. At the end of the matching game, each federated server will have the requested number of clients $C_s$ and each client device will be matched to a federated server. 
\end{enumerate}

The high-level system architecture of our suggested approach is illustrated in Fig.~\ref{fig:Archi}.

\subsection{Matching Fed-IoT Game Formulation}
\label{subsection:matchingametheoryformulation}
In this section, we formulate the Fed-IoT matching game, and establish preference functions of both the federated servers and client IoT devices. We finally provide the algorithms that allow us to create these preference functions. It should be noted that our matching approach is inspired by the methodology discussing \cite{arisdakessian2020fogmatch}.

\textbf{Definition 1}: We define $\gamma$ as a matching relation produced by the matching game between the IoT devices and federated servers, where $\gamma$ is a function $I \cup S \rightarrow 2^{I \cup S}$ that satisfies the following conditions:
\begin{itemize}
    \item $\gamma(i) \subseteq S$, where $|\gamma(i)|= 0$ implies that client $i$ is not assigned to any federated server.
    \item $\gamma(s) \subseteq I$, where that $N_{s} < C_s$ implies that the federated server $s$ didn't reach the needed number of clients for its FL task.
    \item $i \in \gamma(s)$ $if \Leftrightarrow \gamma(i) = s, \forall_i \in I, s \in S$
\end{itemize}

\textbf{Definition 2}: An IoT-Server pair (i, s) is said to block a matching relation $\gamma$ if $\exists$ (i, s) where i $\in \gamma(s)$ and s $\in \gamma(i)$ we have $i \succ_s \gamma(s)$ and $s \succ_i \gamma(i)$.

\textbf{Definition 3}: When a federated server $s$ reaches the needed number of clients $C_s$, it is considered as saturated. If a server still needs some clients, any IoT device $i$ will be accepted as long as it meets the $N_s < C_s$ requirements.

\textbf{Definition 4}: A stable matching relation $\gamma$ exists when (1) there are no blocking relationships, and (2) every federated server is matched with the needed number of client devices.

\subsection{Client IoT Device Preference Function}
\label{IoTMatching}
IoT devices wish to be matched with those federated servers that maximize the former's reward. There exist a complete, strict and transitive preference relation $P_i(S)$ for each IoT device $i \in I$ with each federated server $s \in S$. A preference relationship $s \succ_i s^\prime$ means that federated server $s$ is preferred over federated server $s^\prime$ to IoT device $i$. Furthermore, if an IoT device $i$ does not have a clear preference between joining $s$ or staying unpaired, a federated server $s$ is said to be undesirable to $i$. Based on this description, an IoT device $i$'s preference function can be described as follows: 
    \begin{equation} \label{eq:9}
        s_1 \succ_i s_2 \Leftrightarrow P_i(s_1) > P_i(s_2)
    \end{equation}
where :
\begin{equation} \label{eq:10}
     P_i(s) =
    \begin{cases}
    +\infty,& \quad \text{if $s$ offers the highest reward} \\
    -\infty,& \quad \text{otherwise}
    \end{cases}
\end{equation}

\subsection{Federated Server Preference Function}
\label{ServerMatching}

A federated server prefers to improve the accuracy efficiency by selecting the IoT devices that can train the deep learning model with the best possible local accuracy. There exist a complete, strict and transitive preference relation $P_f(I)$ for each IoT device $s \in S$ with each federated server $i \in I$. A preference relationship $i_1 \succ_s i_2$ means that IoT device $i_1$ is preferred by federated server $s$ over IoT device $i_2$. Furthermore, if $s$ does not have a clear preference between selecting $i$ or staying unpaired, the client device $i$ is supposed to be undesirable to $s$. Based on this description, a federated server $s$'s preference function can be represented as follows:
    \begin{equation} \label{eq:11}
        i_1 \succ_s i_2 \Leftrightarrow P_s(i_1) > P_s(i_2)
    \end{equation}
where :
\begin{equation} \label{eq:12}
     P_s(i) =
    \begin{cases}
    +\infty,& \quad \parbox[t]{.2\textwidth}{if the selection of $i$ maximizes the accuracy of the FL task} \\
    -\infty,& \quad \text{otherwise}
    \end{cases}
\end{equation}

\subsection{Client Device Preference List Creation Algorithm}
\label{}
In Algorithm~\ref{iot_pref}, we show how to establish a preference list for each client IoT device.

\begin{algorithm}[ht]
\caption{Establishment of IoT Device Preference List}
\label{iot_pref}
\begin{algorithmic}[1]
\Require {Collection of active federated servers $S$}
\Ensure {Client IoT device $i$ Preference List $P_i$}
\ForEach {$s$ \textbf{in}  $S$}
\State mark $s$ as visited
\If {data type $\varphi_s$ $\in$ $\varphi_i$}
\State{add $s$ to $P_i$}
\EndIf
\EndFor
\State{\textbf{Sort} the federated servers \textbf{in} $P_i$ Using Eq. (\ref{eq:10})}
\end{algorithmic}
\end{algorithm}

The algorithm takes a collection of federated servers $S$ as input and generates the client IoT device $i$ preference list $P_i$. The federated servers are ordered in the preference list by their preference order to $i$. The algorithm starts by visiting each non-visited federated server (Line $1$) and validating the data type (e.g., MNIST dataset) requested by the federated server (Line $3$). If the requested data type is available on $i$, that corresponding federated server $s$ will be added to $i$'s preference list $P_i$ (Line $4$). Finally, Eq. (\ref{eq:10}) is used to compute the preference ordering across the collection of maintained federated servers in $P_i$ (Line $7$). It is worth nothing mentioning that this algorithm is carried out independently by each IoT device $i$ in $I$ separately.

\subsection{Federated Server Preference List Creation Algorithm}
\label{sec:ServerPrefAlgo}
Federated servers are primarily concerned with increasing the accuracy of training their global models by trying to select the IoT devices with the highest possible accuracy levels.
\begin{algorithm}[ht]\caption{Federated Server Preference List Establishment}
\label{fed_pref}
\begin{algorithmic}[1]
\Require {Collection of client IoT devices willing to participate in the federated learning training}
\Ensure {Preference list $P_s$ of federated server $s$}
\ForEach {$i$ \textbf{$\in$}  $I$}
\State mark $i$ as visited
\If {data type $\varphi_s \in \varphi_i$}
\If {$Acc_{i}$ is \textbf{Unknown}}
\State{// Start Bootstrapping Steps}
\State{Server $s$ sends an inquiry message $\beta^{i}_{s}$ to $\eth$}
\State{$\eth$ receives the message}
\State{$\eth$ collects data from $S$}
\State{$\eth$ rewards each server in $S$ using Eq. (\ref{eq:105})}
\State{$\eth$ Updates the DT model \& predict $\widehat{Acc_i}$}
\State{$\eth$ Replies to the inquiry message by  $\widehat{Acc_i}$}
\State{SET $Acc_{i} = \widehat{Acc_i}$}
\EndIf
\State{add $i$ to $P_s$}
\EndIf
\EndFor
\State{\textbf{Sort} $P_s$ Using Eq. (\ref{eq:12}) to rank the IoT devices \textbf{in} $P_s$}
\end{algorithmic}
\end{algorithm}



Algorithm \ref{fed_pref} is executed by each federated server in $s\in S$ to help build their desired preference lists. The algorithm takes as an input a collection of IoT devices that are eager to participate in the training rounds, and returns the preference list $P_s$ for every federated server. The IoT devices are sorted in the preference list by their order of preference by each $s \in S$. The algorithm initially validates the data type and the local accuracy $Acc_i$ availability at each IoT device in $i \in I$ (Line $1-4$). If the underlying device $i$ has the requested data type and $Acc_i$ is Empty, then the entitled federated server is going to initiate a bootstrapping request $\beta^{i}_{s}$ to the central bootstrapping server $\eth$ (Line $6$). The Trust bootstrapping stages defined before in section~\ref{sec:Bootstrapping Overview} are represented by the steps between Line~($5-12$). After the bootstrapping process is finished, the entitled federated server $s$ assigns the predicted accuracy $\widehat{Acc_i}$ by $\eth$ to $i$. Then, $i$ is added to $P_s$ (Line $14$). Finally, our algorithm employs Eq. (\ref{eq:12}) to assist each federated server in determining the appropriate ordering of the maintained IoT devices (Line $17$).

\section{Selection: Federated Server \& Client Selection Algorithms}
\label{matchingselection}
Once the preference lists have been created, the next step is to devise the appropriate algorithms that would perform the actual matching based on these lists. The end result of this stage is client devices being matched with the federated servers. Our solution is highly distributed in the sense that the IoT devices and federated servers connect directly to accomplish the matching without the need for any third-party central entity.

\subsection{Matching Algorithm - IoT devices}
\label{}

In this section, we provide Algorithm \ref{iot_satching_alg}, which is carried out as part of the matching game by every IoT device. The algorithm accepts as input the preference list in terms of federated servers, obtained after executing Algorithm \ref{fed_pref}. The algorithm goes over the preference list of each IoT device at first (Line $2$) and selects the most desired federated server (Line $3$). The IoT device communicates with the server by sending it a work request message (Line $4$) and waiting for a response (Line $5$). If the server responds positively (i.e., the server agrees to be matched with the IoT device), the loop is broken and the underlying IoT device and federated server are matched (Lines $7-8$). Alternatively, if the reply is negative (Line $6$), the server gets pushed to the bottom of the preference list of the Internet-of-Things device.
The algorithm then proceeds to the next highest priority federated server.
The whole process is repeated for each federated learning round (Line $12$). Note that Algorithm \ref{iot_pref} is executed as part of Algorithm \ref{iot_satching_alg} (Line $11$) to get the latest preference lists of the IoT devices, reflecting the latest variations in the environment in terms of updated performance of the federated servers, advent of new servers and clients to the environment, and removal of some existing ones. 

\begin{algorithm}[ht]
\caption{Internet-of-Things Device Selection Algorithm}
\label{iot_satching_alg}
\begin{algorithmic}[1]
\Require {Internet-of-Things device $i$ Preference list $P_i$}
\Ensure {Client IoT device and federated server matching}
\Repeat
\ForEach {$s \in \mathcal P_i$}
\State mark $s$ as visited
\State{Send a request message $request_{i \rightarrow m}$ to $s$}
\State{Wait for a reply $Reply_{m \rightarrow i}$ from $s$}
\If{$Reply_{s \rightarrow i}$ == 'YES'}
    \State{Match $i$ to $s$}
    \State{Break the Loop}
\EndIf
\EndFor
\State{\textbf{Call} Algorithm \ref{iot_pref}}
\Until{No more federated learning rounds}
\end{algorithmic}
\end{algorithm}

\subsection{Matching Algorithm - Federated Server }
\label{}

Algorithm \ref{fog_satching_alg} is implemented by every federated server, with an input consisting of a queue of IoT devices that have submitted a request message to the underlying federated server. The algorithm checks to see if the queue is empty at first (Line $2$). A non-blank queue indicates that the current federated server is still accepting request messages from the IoT devices. If there are requests in the queue, the algorithm pulls a request message and examines whether the total number of clients selected $N_s$ have reached the requested number of clients $C_s$(Line $3$). If not all needed clients have been selected, the federated server sends an accept message to the underlying IoT device (Line $4$). The federated server then increments the total number of clients selected $N_s$ and adds the IoT device to the selected list $A_{N_s}$ (Line $5-6$). However, if the total number of IoT devices selected $N_s$ is equal to the number of requested IoT devices $C_s$, but the IoT device is ranked better than any already selected device $i^\prime$ such that $i^\prime \in A_{N_s}$ in $P_s$ (Line $7$), the server $s$ will break the agreement with $i^\prime$ (Lines $8-9$) and send an \textit{accept} message to $i$ and add it to the selection list $A_{N_s}$ (Lines $10-11$). Yet, if the IoT device is not ranked better than any already selected IoT device in the federated server's preference list, the IoT device receives a \textit{reject} message (Line $13$) and any IoT devices with a lower preference list rank than the rejected IoT device are removed by the federated server
(Line $14$). It is necessary to point out that the entire process is executed for every federated learning round (Line $18$). To constantly get updated versions of the preference lists that reflect the changes that happen in the environment (i.e., performance variation of IoT devices, arrival/removal of some federated servers and/or IoT devices), Algorithm~\ref{fed_pref} is executed as part of Algorithm \ref{fog_satching_alg} (Line $17$).

\begin{algorithm}[ht]
\caption{Selection Algorithm - Federated server}
\label{fog_satching_alg}
\begin{algorithmic}[1]
\Require {Queue $Q_i$ of IoT devices with $request$ messages to $s$}
\Ensure {Pairing IoT devices with a specific server $s$}
\Repeat
\While{$Q_i$ is not Null}
\If{$N_s < C_s$}
    \State{Send accept reply $\rightarrow Reply_{s \rightarrow i}$ ='Yes'}
    \State{add $i$ to selected $A_{N_s}$}
    \State{increment $N_s$ by 1}
\ElsIf{$N_s == C_s$ \textbf{and} $i \succ_s i^\prime $} \Comment{where $i^\prime$ is the worst selected IoT in $A_{N_s}$}
    \State{Send updated rejection $Reply_{s \rightarrow i^\prime}$ = 'No'}
    \State{remove $i^\prime$ from the accepted list $A_{N_s}$}
    \State{Send accept reply $\rightarrow Reply_{s \rightarrow i}$ = 'Yes'}
    \State{add $i$ to the accepted list $A_{N_s}$}
\Else
    \State{Send rejection reply to $i \rightarrow Reply_{s \rightarrow i}$ = 'No'}
    \State{Discard every $i^\prime \in Q_i$ where $i \succ_s i^\prime $}
\EndIf
\EndWhile
\State{\textbf{Call} Algorithm \ref{fed_pref}}
\Until{No more federated learning rounds}
\end{algorithmic}
\end{algorithm}

\section{Experiments}
\label{sec:experiments}

In this section, we describe the context in which we ran our simulations and discuss the findings of our experiments.

\subsection{Experimental Setup}
\label{experimentalsetup}

In our simulations, we use the MNIST dataset from the National Institute of Standards and Technology (NIST)\footnote{http://yann.lecun.com/exdb/mnist/}. The training sample comprises manually written digits by $250$ distinct persons, $50\%$ of them are secondary school students and $50\%$ are from the Census Bureau. The testing dataset, likewise, includes the same distribution of manually written digital information. The MNIST dataset comprises $60k$ pictures as training dataset and $10k$  as testing dataset, all having a size of $28 \times 28$ pixels and $256$ gray levels \cite{deng2012mnist}. In terms of label and size distributions, the dataset was split over the client IoT devices in an non-IID fashion. An initial set of $C = 100$ IoT devices was created and each round 10 clients are going to be added, each having a dataset size in the interval of $[100,450]$ images. Each IoT device has at least one label and no more than four labels in its class label distribution. To run the simulations, we build our own platform in which each IoT device has a CPU capacity ranged between $300$ and $700$ MIPS, RAM capacity ranged between $400$ and $900$ MB, and bandwidth ranged between $500$ and $900$ Mbps. The latency across each federated server and IoT device pair varies between $0.1$ and $5$ seconds.

We compare our solution with the baseline \textit{VanillaFL} method which was first introduced by Google \cite{mcmahan2017communication}, where the client selection happens randomly. For our simulations, we start with initial set $C$ as $100$ and add 10 clients each round. We set $K$ to $10$ for the \textit{VanillaFL} same value apply to the $C_s$ in the \textit{FedMint}, number of federated servers to $2$, and total number of FL rounds $R$ to $15$.

\subsection{Experimental results}

We discuss the findings of our experiments in two subsections: (1) FedMint vs VanillaFL and (2) Bootstrapping results.
\label{experimentalresults}

\subsubsection{FedMint vs VanillaFL}
Our experiments are primarily designed to compare \textit{FedMint} approach versus \textit{VanillaFL} by investigating two key metrics. First, IoT monetary rewards and Second, global model accuracy. Then we study the results of applying the Bootstrapping versus random accuracy in the Global model Accuracy context. 

\begin{figure}[ht]
\subfloat[Federated Server 1 (proposed approach vs VanillaFL)]{\includegraphics[width=\linewidth]{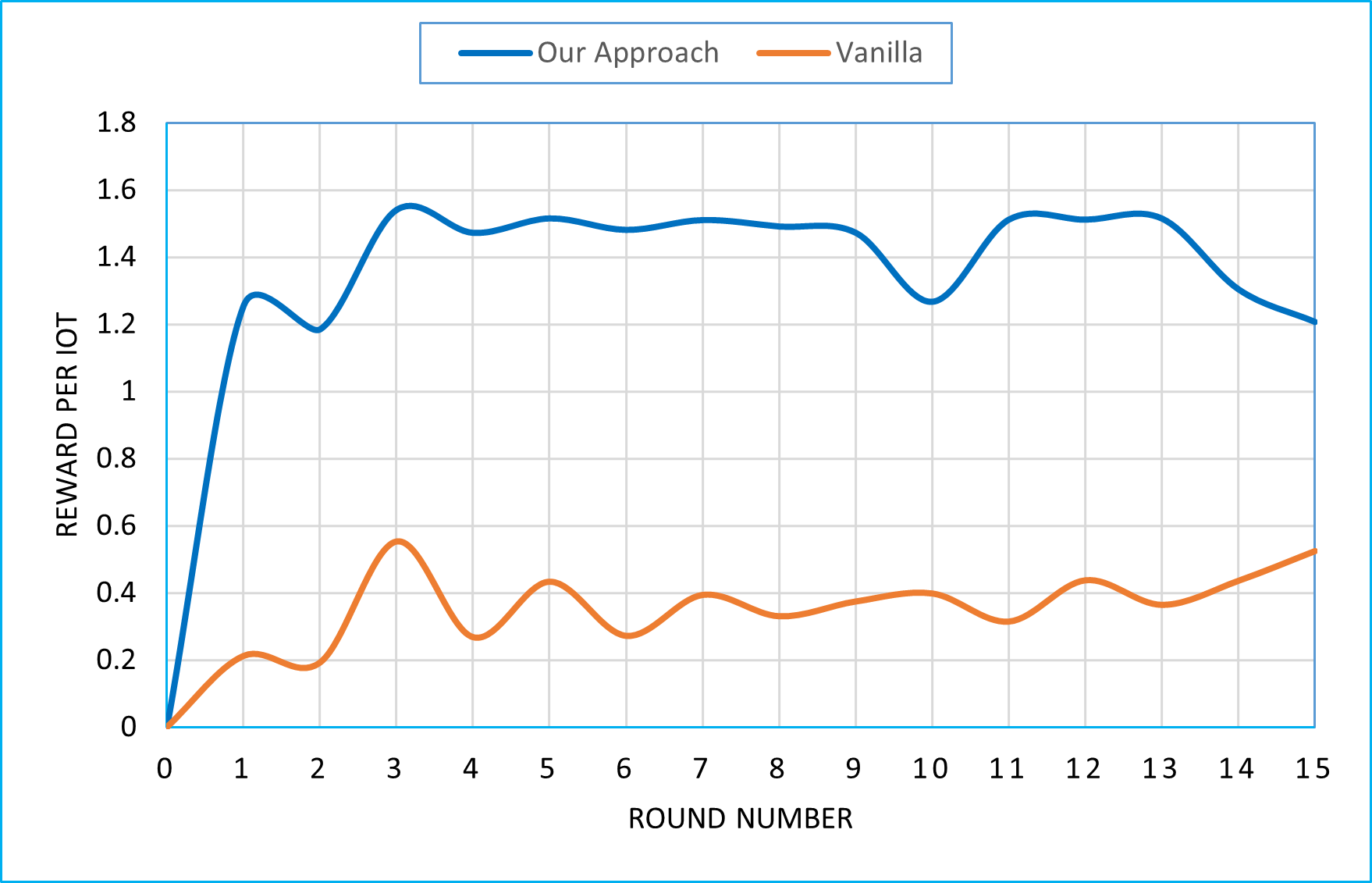}}
\hfill
\subfloat[Federated Server 2 (proposed approach vs VanillaFL)]{\includegraphics[width=\linewidth]{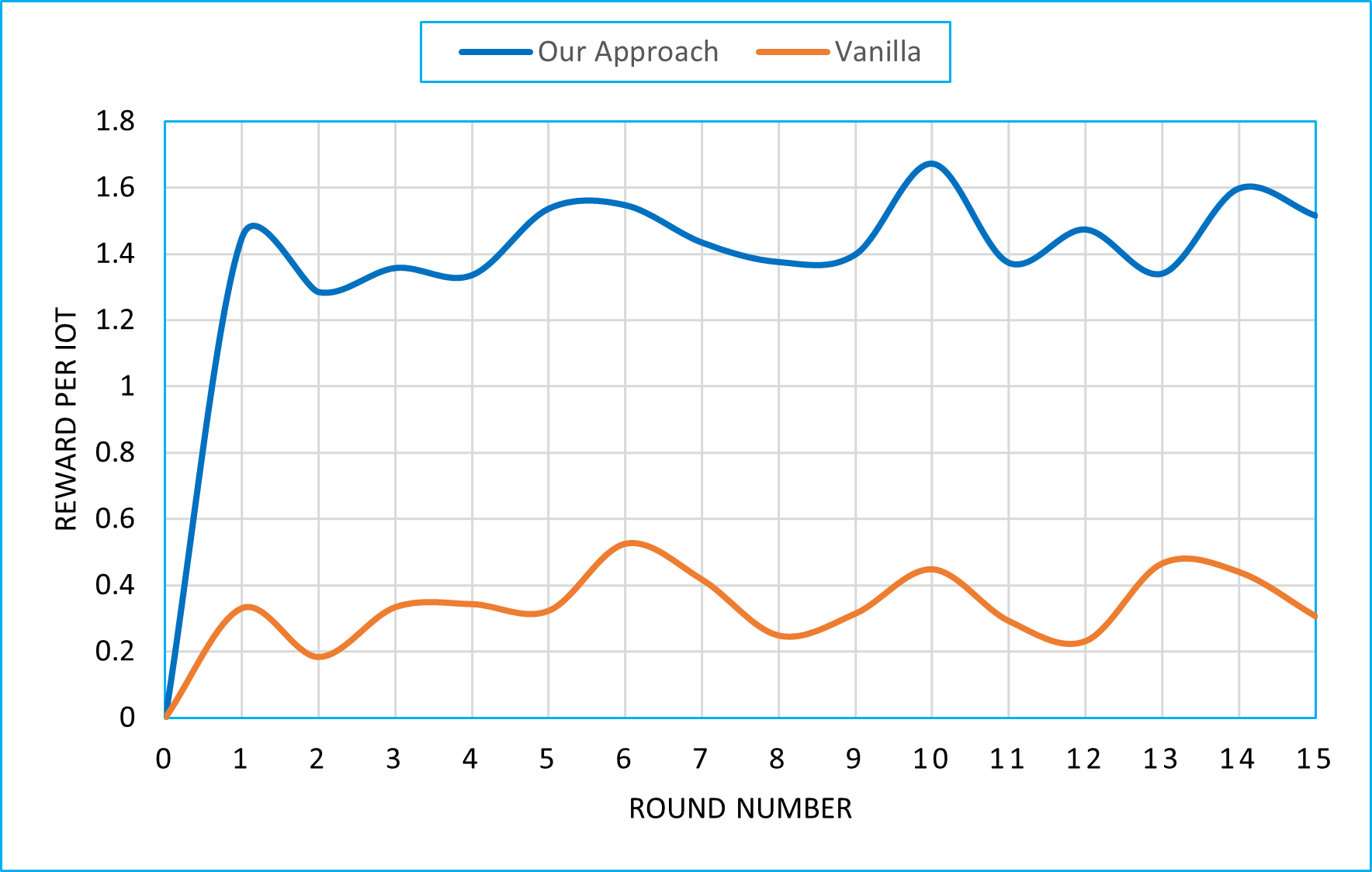}}
\caption{Our solution increases the IoT devices rewards compared to~\cite{mcmahan2017communication} by more than $58\%$.}
\label{fig:res2}
\end{figure}   

\begin{itemize}
    \item \textbf{IoT Devices Monetary Rewards:} In Fig.~\ref{fig:res2}, we study the average reward that an IoT device gains after having participated in the federated learning task, over the federated learning rounds number.
    Fig.~\ref{fig:res2} shows that our solution maximizes the clients rewards considerable compared to \textit{VanillaFL}. This stems from the fact that our solution takes into consideration each client's preference in the selection mechanism. On the other hand, the random client selection in VanillaFL is server-oriented, thus totally ignoring the client's preferences in the selection process. We notice from Fig.~\ref{fig:res2} that (a) the rewards of the IoT devices that participated in federated learning rounds with Federated Server $1$ using our approach are higher than those obtained by the IoT devices in \textit{VanillaFL} in all the FL rounds by $58.29\%$ with minimum and maximum differences of $37.31\%$ and $67.19\%$ respectively.
    Similarly, in the case of Federated Server 2 (Fig.~\ref{fig:res2} (b)), our solution enables the IoT devices to obtain higher rewards than those obtained in \textit{VanillaFL} in average by $61.14\%$ inline with minimum and maximum differences of $48.67\%$ and $69.11\%$ respectively.\\

    \begin{figure}[ht]
    \subfloat[Federated Server 1 (proposed approach vs VanillaFL)]
    {\includegraphics[width=\linewidth]{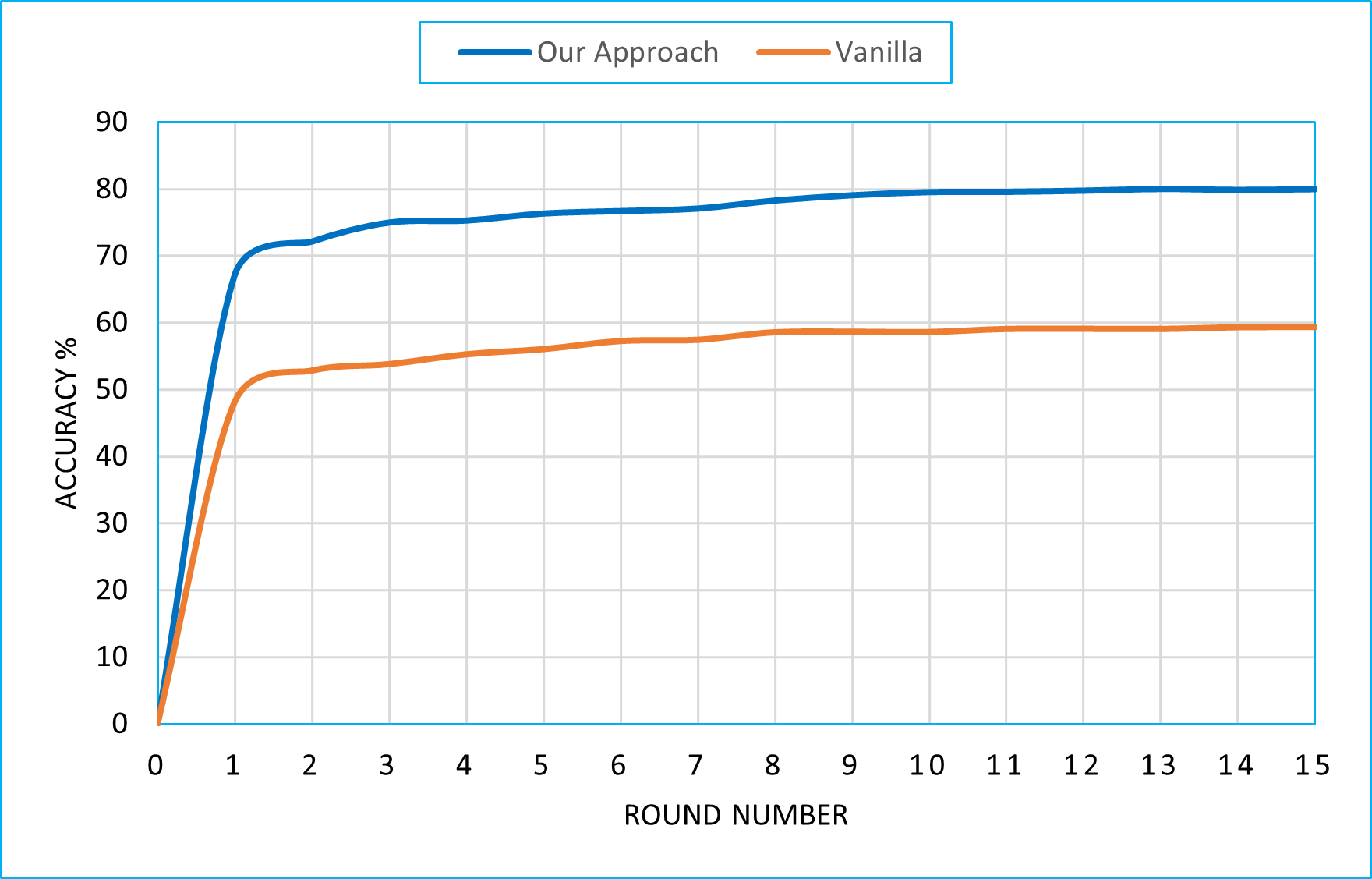}}
    \hfill
    \subfloat[Federated Server 2 (proposed approach vs VanillaFL)]{\includegraphics[width=\linewidth]{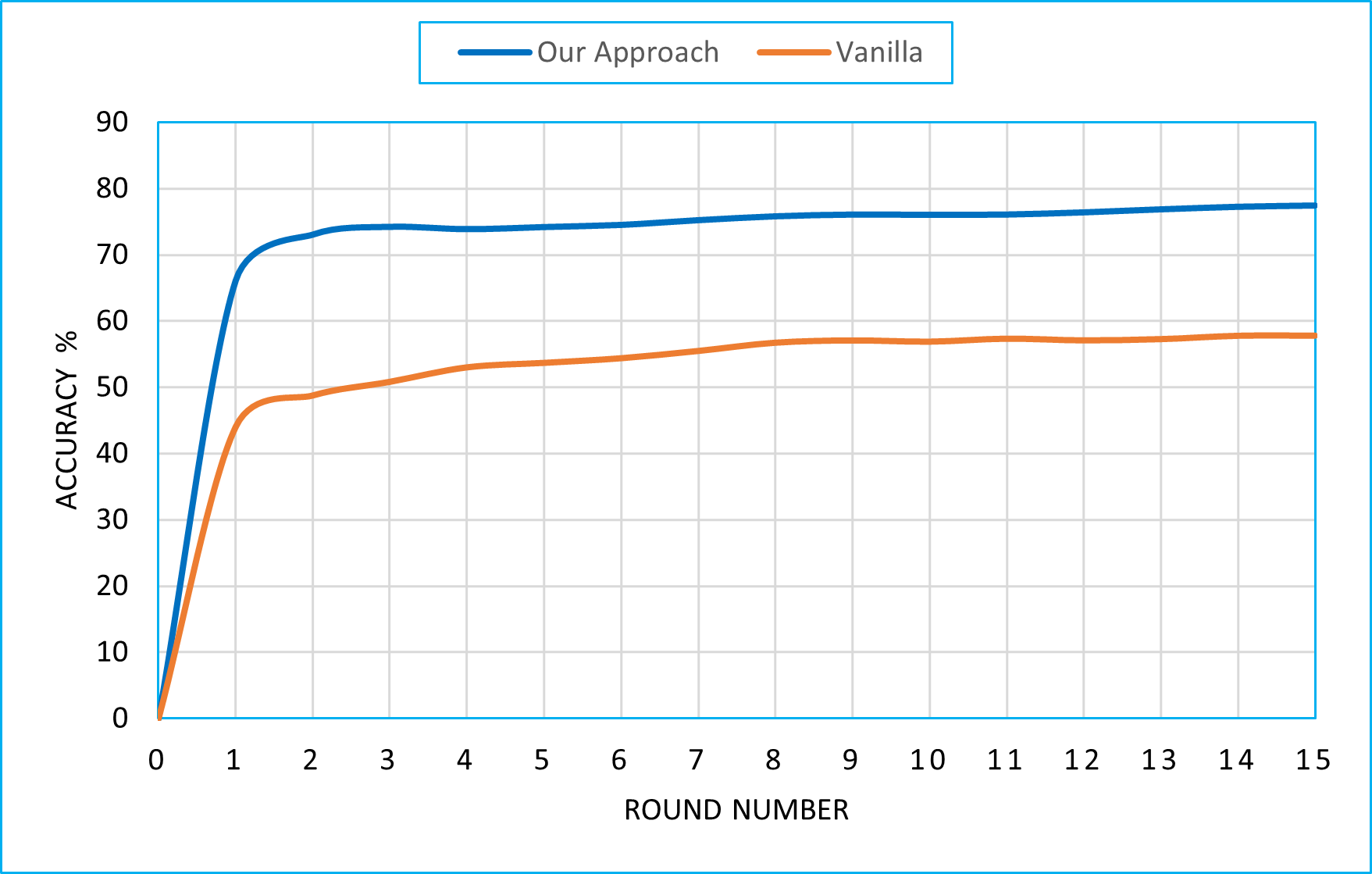}}
    \caption{Our solution achieves high accuracy levels compared to~\cite{mcmahan2017communication} by more than $20\%$.}
    \label{fig:res1}
    \end{figure}

    \item \textbf{Federated Learning Model Accuracy:} Each federated server aims to maximize its global model's accuracy by selecting the IoT devices with the highest local accuracy. In Fig.~\ref{fig:res1}, We measure the global model's accuracy in both our solution and \textit{VanillaFL}. The results illustrated in Fig.~\ref{fig:res1} (a, b) represent the average global accuracy for Federated Server $1$ and Federated Server $2$ respectively, with respect to the FL communication rounds. To avoid biased results, we run each method several times and compute the average accuracy. Overall, as can be seen in Fig.~\ref{fig:res1}(a), our solution outperforms \textit{VanillaFL} in terms of model's accuracy. Specifically, our solution achieves an accuracy level of $66.12\%$ in the first communication round compared to $43.99\%$ in \textit{VanillaFL} in the case of Federated Server $1$. Similarly, we notice from Fig.~\ref{fig:res1}(b) the accuracy of our solution in the first communication round is $67.10\%$ compared to $48.18\%$ in \textit{VanillaFL}. In the $15th$ rounds, both our solution and \textit{VanillaFL} reach their highest accuracy levels on both servers. In more detail, Federated Server $1$ reaches an accuracy level of $77.45\%$ which is considerably higher than that of \textit{VanillaFL} by $19.70\%$. Federated Server $2$ reaches its highest accuracy level of $79.92\%$ which is higher than that of \textit{VanillaFL} by $20.56\%$. 

\end{itemize}


\subsubsection{Bootstrapping results}
\begin{figure}[ht]
    \centering
    \includegraphics[width=\linewidth]{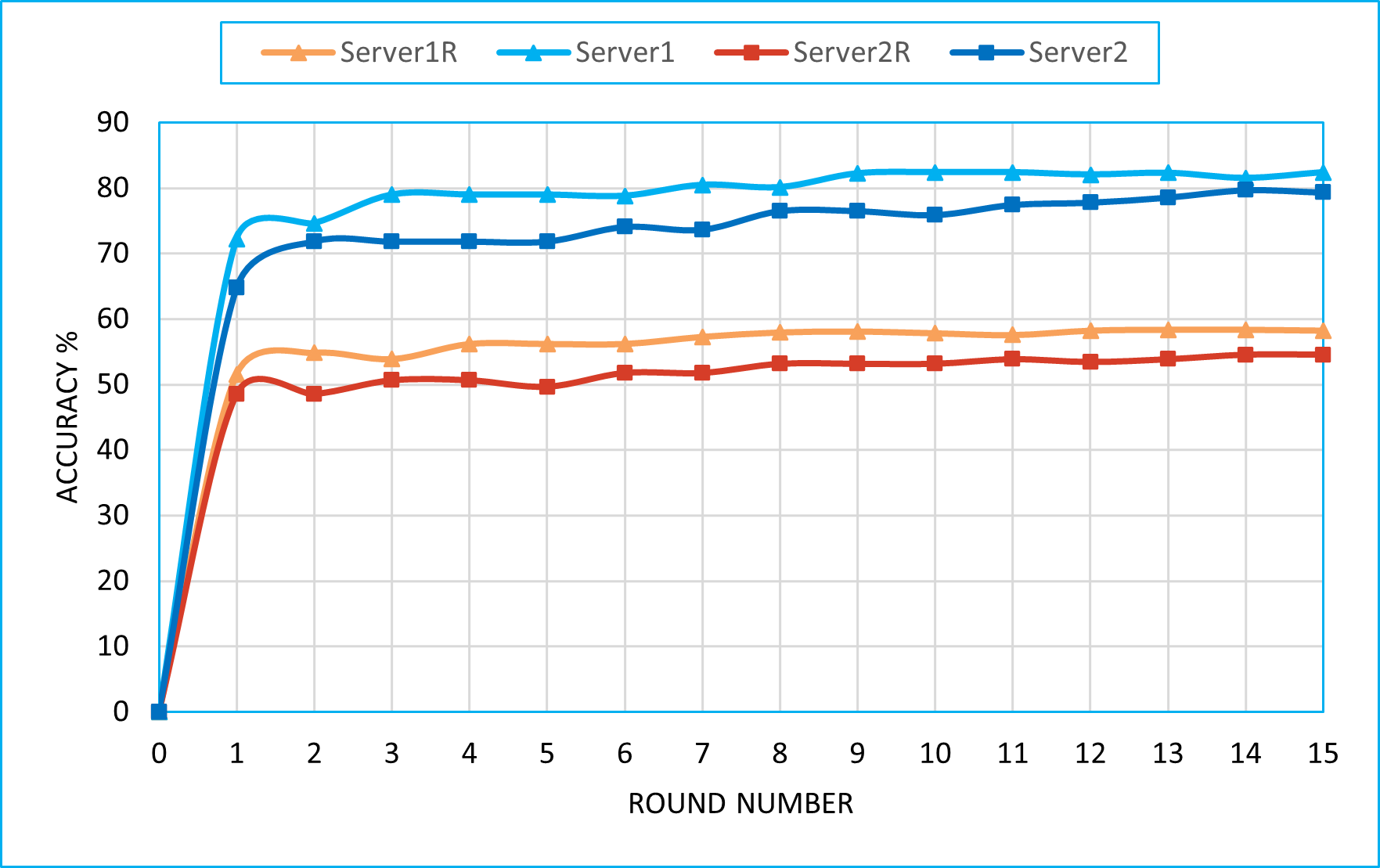}
    \caption{Global model accuracy rate of the FedMint approach using bootstrapping vs randomly assigned accuracy rate with respect to communication round.}
    \label{fig:Fedboot}
\end{figure}

\begin{figure}[ht]
    \centering
    \includegraphics[width=\linewidth]{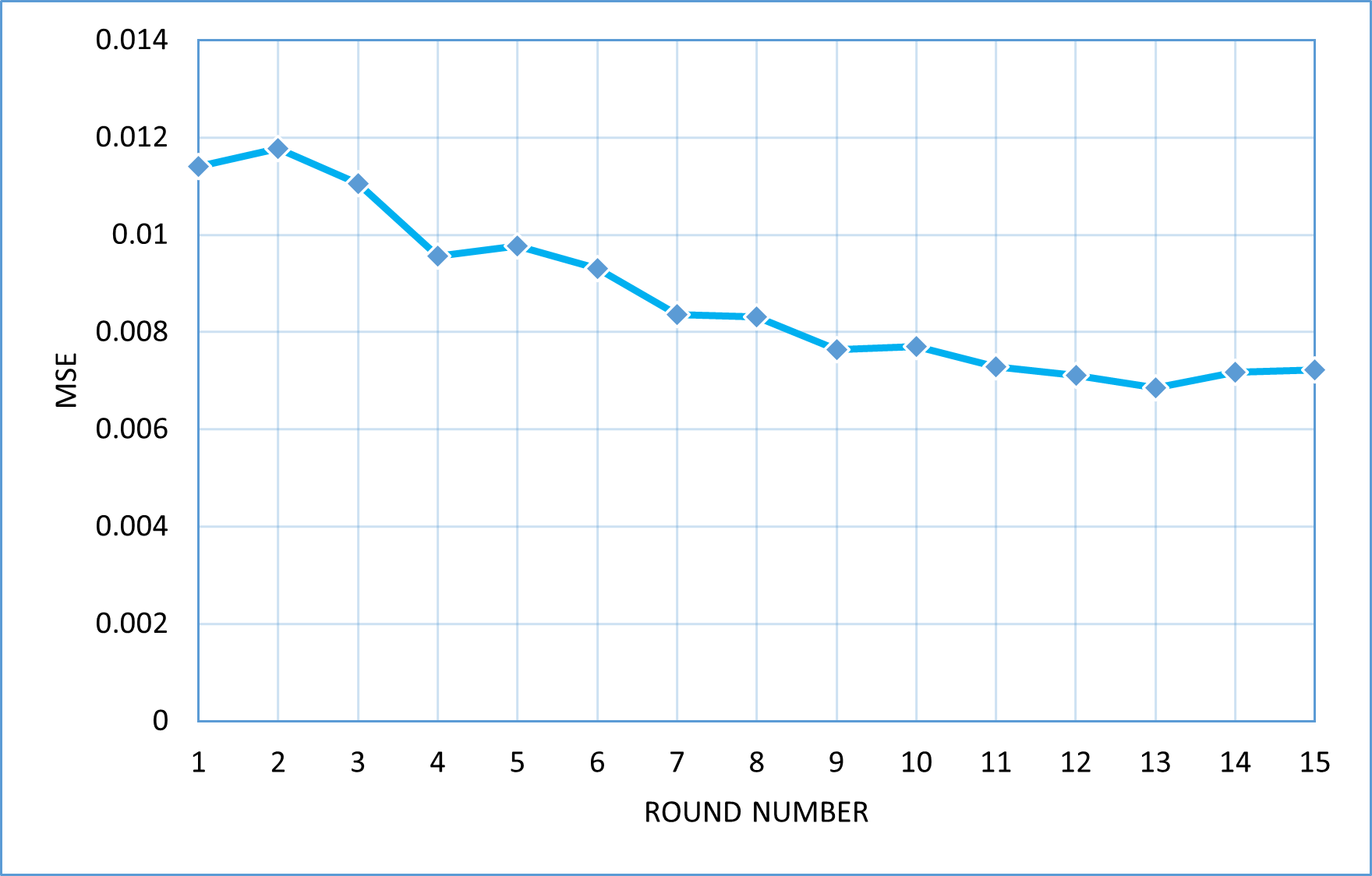}
    \caption{Mean Square Error performance with respect to communication round using K=10 K-fold cross validation.}
    \label{fig:MSE}
\end{figure}

We investigate the trust bootstrapping mechanism efficiency by showing the accuracy results for the \textit{FedMint} approach in Fig.~(\ref{fig:Fedboot}) by assigning random initial accuracy versus using bootstrapping for getting initial accuracy for the newcomer IoT devices. The MSE plot in Fig.~(\ref{fig:MSE}) for the Bootstrapping decision tree model is then interpreted. The MSE measures the amount of error in machine learning models. For a perfect model the MSE value is $0$ and this value increases as the model error increases.

As shown in Fig.~(\ref{fig:Fedboot}), the accuracy obtained by applying bootstrapping illustrated by the two upper lines represent Federated Servers $1$ and $2$ respectively is quite better. In case of Federated Server $1$, the light-blue upper line represents the results using our bootstrapping. On the other hand, the lower light-green line represents the results by assigning random accuracy, we notice that bootstrapping solution achieves an accuracy level of $72.21\%$ in the first communication round compared to $51.56\%$ in the random accuracy approach. Similarly, for Federated Server $2$ illustrated by the dark-blue line for the federated server that applies the bootstrapping for initial accuracy versus the red line that use randomly assigned accuracy. We notice that the bootstrapping solution achieves an accuracy level of $64.77\%$ in the first communication round compared to $48.53\%$ in the random accuracy approach. In the $15th$ round, both Federated Servers $1$ and $2$ that adopt our bootstrapping solution achieve their highest accuracy represented by $82.38\%$ and $79.32\%$ respectively, which is higher than the random accuracy approach by more than $24\%$.

In Fig.~(\ref{fig:MSE}), we study the performance of the trust bootstrapping Decision Tree model by applying K-Fold cross validation to plot the model's MSE with K=10.
Fig.~(\ref{fig:MSE}) shows that our model performs very well since the maximum MSE value reached is very low around $0.0118$. Also, we observe from the figure that the MSE value decreases as the round number increases which mean that the bootstrapping model performance is improving, where the MSE value at the first round was $0.0115$ which is less by $0.0042$ than the last round equal to $0.0073$.

\section{Conclusion}
\label{Conclusion}

In this research, we introduced \textit{FedMint}, a bilateral matching approach with newcomer IoT devices. The proposed solution takes into consideration the preferences and constraints of both the federated servers and client devices in its design, along with a trust bootstrapping system for newcomer IoT devices initial accuracy assignments. Simulation results indicate that our solution maximizes the federated learning global model accuracy from the perspective of the federated servers, and the monetary rewards from the perspective of the client IoT devices. We compared our method performance to that of the baseline random client selection FL strategy. Our findings suggest that, our approach improves accuracy by more than 20\%, as well as our solution boosts income of the participating IoT devices by more than 58\%.

\bibliographystyle{IEEEtran}
\bibliography{references}

\begin{thebibliography}{10}
\providecommand{\url}[1]{#1}
\csname url@samestyle\endcsname
\providecommand{\newblock}{\relax}
\providecommand{\bibinfo}[2]{#2}
\providecommand{\BIBentrySTDinterwordspacing}{\spaceskip=0pt\relax}
\providecommand{\BIBentryALTinterwordstretchfactor}{4}
\providecommand{\BIBentryALTinterwordspacing}{\spaceskip=\fontdimen2\font plus
\BIBentryALTinterwordstretchfactor\fontdimen3\font minus
  \fontdimen4\font\relax}
\providecommand{\BIBforeignlanguage}[2]{{%
\expandafter\ifx\csname l@#1\endcsname\relax
\typeout{** WARNING: IEEEtran.bst: No hyphenation pattern has been}%
\typeout{** loaded for the language `#1'. Using the pattern for}%
\typeout{** the default language instead.}%
\else
\language=\csname l@#1\endcsname
\fi
#2}}
\providecommand{\BIBdecl}{\relax}
\BIBdecl

\bibitem{wahab2022intrusion}
O.~A. Wahab, ``Intrusion detection in the iot under data and concept drifts:
  Online deep learning approach,'' \emph{IEEE Internet of Things Journal},
  2022.

\bibitem{wang2021toward}
X.~Wang, S.~Garg, H.~Lin, J.~Hu, G.~Kaddoum, M.~J. Piran, and M.~S. Hossain,
  ``Toward accurate anomaly detection in industrial internet of things using
  hierarchical federated learning,'' \emph{IEEE Internet of Things Journal},
  vol.~9, no.~10, pp. 7110--7119, 2021.

\bibitem{nguyen2021federated}
D.~C. Nguyen, M.~Ding, P.~N. Pathirana, A.~Seneviratne, J.~Li, and H.~V. Poor,
  ``Federated learning for internet of things: A comprehensive survey,''
  \emph{IEEE Communications Surveys \& Tutorials}, 2021.

\bibitem{elayan2021sustainability}
H.~Elayan, M.~Aloqaily, and M.~Guizani, ``Sustainability of healthcare data
  analysis iot-based systems using deep federated learning,'' \emph{IEEE
  Internet of Things Journal}, vol.~9, no.~10, pp. 7338--7346, 2021.

\bibitem{sorkhoh2022optimizing}
I.~Sorkhoh, M.~A. Arfaoui, M.~Khabbaz, and C.~Assi, ``Optimizing information
  freshness in ris-assisted cooperative autonomous driving,'' in \emph{ICC
  2022-IEEE International Conference on Communications}.\hskip 1em plus 0.5em
  minus 0.4em\relax IEEE, 2022, pp. 1518--1523.

\bibitem{index2016forecast}
C.~V.~N. Index, ``Forecast and methodology, 2015--2020,'' \emph{White paper},
  pp. 1--41, 2016.

\bibitem{pei2022knowledge}
X.~Pei, X.~Deng, S.~Tian, L.~Zhang, and K.~Xue, ``A knowledge transfer-based
  semi-supervised federated learning for iot malware detection,'' \emph{IEEE
  Transactions on Dependable and Secure Computing}, 2022.

\bibitem{li2021resource}
R.~Li, P.~Hong, K.~Xue, M.~Zhang, and T.~Yang, ``Resource allocation for uplink
  noma-based d2d communication in energy harvesting scenario: A two-stage game
  approach,'' \emph{IEEE Transactions on Wireless Communications}, vol.~21,
  no.~2, pp. 976--990, 2021.

\bibitem{sorkhoh2020infrastructure}
I.~Sorkhoh, D.~Ebrahimi, C.~Assi, S.~Sharafeddine, and M.~Khabbaz, ``An
  infrastructure-assisted workload scheduling for computational resources
  exploitation in the fog-enabled vehicular network,'' \emph{IEEE Internet of
  Things Journal}, vol.~7, no.~6, pp. 5021--5032, 2020.

\bibitem{rathee2022trustsys}
G.~Rathee, S.~Garg, G.~Kaddoum, B.~J. Choi, M.~Hassan, and S.~A. Alqahtani,
  ``Trustsys: Trusted decision making scheme for collaborative artificial
  intelligence of things,'' \emph{IEEE Transactions on Industrial Informatics},
  2022.

\bibitem{rasheed2022explainable}
K.~Rasheed, A.~Qayyum, M.~Ghaly, A.~Al-Fuqaha, A.~Razi, and J.~Qadir,
  ``Explainable, trustworthy, and ethical machine learning for healthcare: A
  survey,'' \emph{Computers in Biology and Medicine}, p. 106043, 2022.

\bibitem{wazzeh2022privacy}
M.~Wazzeh, H.~Ould-Slimane, C.~Talhi, A.~Mourad, and M.~Guizani,
  ``Privacy-preserving continuous authentication for mobile and iot systems
  using warmup-based federated learning,'' \emph{IEEE Network}, 2022.

\bibitem{otoum2022federated}
S.~Otoum, I.~Al~Ridhawi, and H.~Mouftah, ``A federated learning and
  blockchain-enabled sustainable energy-trade at the edge: A framework for
  industry 4.0,'' \emph{IEEE Internet of Things Journal}, 2022.

\bibitem{rahman2020internet}
S.~A. Rahman, H.~Tout, C.~Talhi, and A.~Mourad, ``Internet of things intrusion
  detection: Centralized, on-device, or federated learning?'' \emph{IEEE
  Network}, vol.~34, no.~6, pp. 310--317, 2020.

\bibitem{tabassum2022fedgan}
A.~Tabassum, A.~Erbad, W.~Lebda, A.~Mohamed, and M.~Guizani, ``Fedgan-ids:
  Privacy-preserving ids using gan and federated learning,'' \emph{Computer
  Communications}, vol. 192, pp. 299--310, 2022.

\bibitem{rjoub2022trust}
G.~Rjoub, O.~A. Wahab, J.~Bentahar, and A.~Bataineh, ``Trust-driven
  reinforcement selection strategy for federated learning on iot devices,''
  \emph{Computing}, pp. 1--23, 2022.

\bibitem{konevcny2016federated}
J.~Kone{\v{c}}n{\`y}, H.~B. McMahan, F.~X. Yu, P.~Richt{\'a}rik, A.~T. Suresh,
  and D.~Bacon, ``Federated learning: Strategies for improving communication
  efficiency,'' \emph{arXiv preprint arXiv:1610.05492}, 2016.

\bibitem{wahab2021federated}
O.~A. Wahab, A.~Mourad, H.~Otrok, and T.~Taleb, ``Federated machine learning:
  Survey, multi-level classification, desirable criteria and future directions
  in communication and networking systems,'' \emph{IEEE Communications Surveys
  \& Tutorials}, vol.~23, no.~2, pp. 1342--1397, 2021.

\bibitem{qayyum2022collaborative}
A.~Qayyum, K.~Ahmad, M.~A. Ahsan, A.~Al-Fuqaha, and J.~Qadir, ``Collaborative
  federated learning for healthcare: Multi-modal covid-19 diagnosis at the
  edge,'' \emph{IEEE Open Journal of the Computer Society}, 2022.

\bibitem{abdulrahman2020survey}
S.~AbdulRahman, H.~Tout, H.~Ould-Slimane, A.~Mourad, C.~Talhi, and M.~Guizani,
  ``A survey on federated learning: The journey from centralized to distributed
  on-site learning and beyond,'' \emph{IEEE Internet of Things Journal},
  vol.~8, no.~7, pp. 5476--5497, 2020.

\bibitem{kairouz2021advances}
P.~Kairouz, H.~B. McMahan, B.~Avent, A.~Bellet, M.~Bennis, A.~N. Bhagoji,
  K.~Bonawitz, Z.~Charles, G.~Cormode, R.~Cummings \emph{et~al.}, ``Advances
  and open problems in federated learning,'' \emph{Foundations and
  Trends{\textregistered} in Machine Learning}, vol.~14, no. 1--2, pp. 1--210,
  2021.

\bibitem{mcmahan2017communication}
B.~McMahan, E.~Moore, D.~Ramage, S.~Hampson, and B.~A. y~Arcas,
  ``Communication-efficient learning of deep networks from decentralized
  data,'' in \emph{Artificial intelligence and statistics}.\hskip 1em plus
  0.5em minus 0.4em\relax PMLR, 2017, pp. 1273--1282.

\bibitem{albaseer2022balanced}
A.~Albaseer, M.~Abdallah, A.~Al-Fuqaha, and A.~Erbad, ``Balanced energy
  consumption based on historical participation of resource-constrained devices
  in federated edge learning,'' in \emph{2022 International Wireless
  Communications and Mobile Computing (IWCMC)}.\hskip 1em plus 0.5em minus
  0.4em\relax IEEE, 2022, pp. 300--305.

\bibitem{hammoud2022demand}
A.~Hammoud, H.~Otrok, A.~Mourad, and Z.~Dziong, ``On demand fog federations for
  horizontal federated learning in iov,'' \emph{IEEE Transactions on Network
  and Service Management}, 2022.

\bibitem{hammoud2022data}
A.~Hammoud, A.~Mourad, H.~Otrok, and Z.~Dziong, ``Data-driven federated
  autonomous driving,'' in \emph{International Conference on Mobile Web and
  Intelligent Information Systems}.\hskip 1em plus 0.5em minus 0.4em\relax
  Springer, 2022, pp. 79--90.

\bibitem{chiti2018matching}
F.~Chiti, R.~Fantacci, and B.~Picano, ``A matching theory framework for tasks
  offloading in fog computing for iot systems,'' \emph{IEEE Internet of Things
  Journal}, vol.~5, no.~6, pp. 5089--5096, 2018.

\bibitem{roth1992two}
A.~E. Roth and M.~Sotomayor, ``Two-sided matching,'' \emph{Handbook of game
  theory with economic applications}, vol.~1, pp. 485--541, 1992.

\bibitem{arisdakessian2022survey}
S.~Arisdakessian, O.~A. Wahab, A.~Mourad, H.~Otrok, and M.~Guizani, ``A survey
  on iot intrusion detection: Federated learning, game theory, social
  psychology and explainable ai as future directions,'' \emph{IEEE Internet of
  Things Journal}, 2022.

\bibitem{chiti2020matching}
F.~Chiti, R.~Fantacci, and B.~Picano, ``A matching game for tasks offloading in
  integrated edge-fog computing systems,'' \emph{Transactions on Emerging
  Telecommunications Technologies}, vol.~31, no.~2, p. e3718, 2020.

\bibitem{goetz2019active}
J.~Goetz, K.~Malik, D.~Bui, S.~Moon, H.~Liu, and A.~Kumar, ``Active federated
  learning,'' \emph{arXiv preprint arXiv:1909.12641}, 2019.

\bibitem{cho2020client}
Y.~J. Cho, J.~Wang, and G.~Joshi, ``Client selection in federated learning:
  Convergence analysis and power-of-choice selection strategies,'' \emph{arXiv
  preprint arXiv:2010.01243}, 2020.

\bibitem{nishio2019client}
T.~Nishio and R.~Yonetani, ``Client selection for federated learning with
  heterogeneous resources in mobile edge,'' in \emph{ICC 2019-2019 IEEE
  international conference on communications (ICC)}.\hskip 1em plus 0.5em minus
  0.4em\relax IEEE, 2019, pp. 1--7.

\bibitem{abdulrahman2020fedmccs}
S.~AbdulRahman, H.~Tout, A.~Mourad, and C.~Talhi, ``Fedmccs: Multicriteria
  client selection model for optimal iot federated learning,'' \emph{IEEE
  Internet of Things Journal}, vol.~8, no.~6, pp. 4723--4735, 2020.

\bibitem{huang2020efficiency}
T.~Huang, W.~Lin, W.~Wu, L.~He, K.~Li, and A.~Y. Zomaya, ``An
  efficiency-boosting client selection scheme for federated learning with
  fairness guarantee,'' \emph{IEEE Transactions on Parallel and Distributed
  Systems}, vol.~32, no.~7, pp. 1552--1564, 2020.

\bibitem{zhao2022participant}
J.~Zhao, X.~Chang, Y.~Feng, C.~H. Liu, and N.~Liu, ``Participant selection for
  federated learning with heterogeneous data in intelligent transport system,''
  \emph{IEEE Transactions on Intelligent Transportation Systems}, 2022.

\bibitem{chen2021matching}
D.~Chen, C.~S. Hong, L.~Wang, Y.~Zha, Y.~Zhang, X.~Liu, and Z.~Han,
  ``Matching-theory-based low-latency scheme for multitask federated learning
  in mec networks,'' \emph{IEEE Internet of Things Journal}, vol.~8, no.~14,
  pp. 11\,415--11\,426, 2021.

\bibitem{kang2020training}
J.~Kang, Z.~Xiong, D.~Niyato, Z.~Cao, and A.~Leshem, ``Training task allocation
  in federated edge learning: A matching-theoretic approach,'' in \emph{2020
  IEEE 17th Annual Consumer Communications \& Networking Conference
  (CCNC)}.\hskip 1em plus 0.5em minus 0.4em\relax IEEE, 2020, pp. 1--6.

\bibitem{cao2020fltrust}
X.~Cao, M.~Fang, J.~Liu, and N.~Z. Gong, ``Fltrust: Byzantine-robust federated
  learning via trust bootstrapping,'' \emph{arXiv preprint arXiv:2012.13995},
  2020.

\bibitem{dong2021flod}
Y.~Dong, X.~Chen, K.~Li, D.~Wang, and S.~Zeng, ``Flod: Oblivious defender for
  private byzantine-robust federated learning with dishonest-majority,'' in
  \emph{Computer Security--ESORICS 2021: 26th European Symposium on Research in
  Computer Security, Darmstadt, Germany, October 4--8, 2021, Proceedings, Part
  I}, 2021, pp. 497--518.

\bibitem{wahab2022federated}
O.~A. Wahab, G.~Rjoub, J.~Bentahar, and R.~Cohen, ``Federated against the cold:
  A trust-based federated learning approach to counter the cold start problem
  in recommendation systems,'' \emph{Information Sciences}, vol. 601, pp.
  189--206, 2022.

\bibitem{wahab2020endorsement}
O.~A. Wahab, R.~Cohen, J.~Bentahar, H.~Otrok, A.~Mourad, and G.~Rjoub, ``An
  endorsement-based trust bootstrapping approach for newcomer cloud services,''
  \emph{Information Sciences}, vol. 527, pp. 159--175, 2020.

\bibitem{wahab2016towards}
O.~A. Wahab, J.~Bentahar, H.~Otrok, and A.~Mourad, ``Towards trustworthy
  multi-cloud services communities: A trust-based hedonic coalitional game,''
  \emph{IEEE Transactions on Services Computing}, vol.~11, no.~1, pp. 184--201,
  2016.

\bibitem{quinlan1986induction}
J.~R. Quinlan, ``Induction of decision trees,'' \emph{Machine learning},
  vol.~1, no.~1, pp. 81--106, 1986.

\bibitem{quinlan1992learning}
J.~R. Quinlan \emph{et~al.}, ``Learning with continuous classes,'' in \emph{5th
  Australian joint conference on artificial intelligence}, vol.~92.\hskip 1em
  plus 0.5em minus 0.4em\relax World Scientific, 1992, pp. 343--348.

\bibitem{arisdakessian2020fogmatch}
S.~Arisdakessian, O.~A. Wahab, A.~Mourad, H.~Otrok, and N.~Kara, ``Fogmatch: an
  intelligent multi-criteria iot-fog scheduling approach using game theory,''
  \emph{IEEE/ACM Transactions on Networking}, vol.~28, no.~4, pp. 1779--1789,
  2020.

\bibitem{deng2012mnist}
L.~Deng, ``The mnist database of handwritten digit images for machine learning
  research,'' \emph{IEEE Signal Processing Magazine}, vol.~29, no.~6, pp.
  141--142, 2012.

\end{thebibliography}

\begin{IEEEbiography}[{\includegraphics[width=1in,height=1.25in,clip,keepaspectratio]{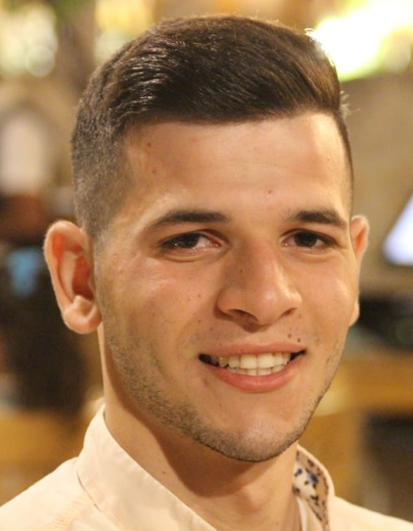}}]%
{\textbf{\textit{\textbf{Osama Wehbi}}}}
received his Master One in business computing from the Lebanese University in 2020 and M.Sc in Computer Science from the Lebanese American University in 2022. He was a Research and Teaching assistant at the Lebanese American University. His current research interests include Federated Machine Learning, Game Theory, Federated Cloud and Fog, Artificial Intelligence, and Cyber Security. 
\end{IEEEbiography}

\vskip 0pt plus -1fil

\begin{IEEEbiography}[{\includegraphics[width=1in,height=1.25in,clip,keepaspectratio]{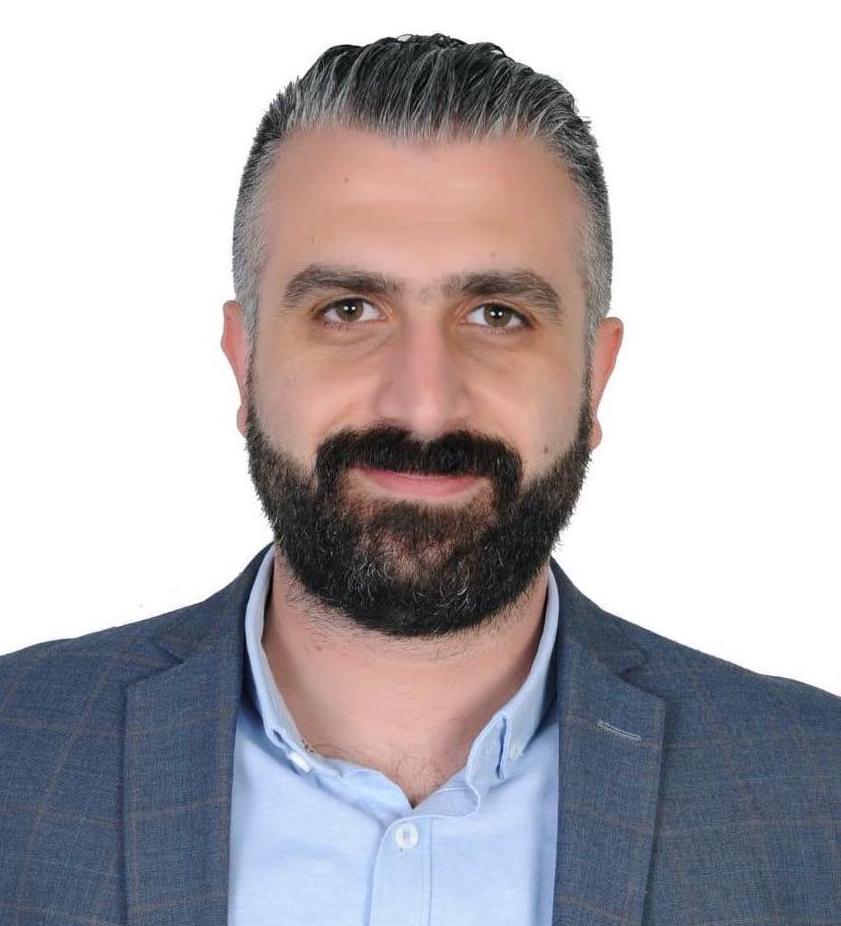}}]%
{\textbf{\textit{\textbf{Sarhad Arisdakessian}}}}
is a PhD student in Computer Science in the Department of Computer Engineering at Polytechnique Montréal, Canada. He had his Master's degree in Computer Science from the Lebanese American University (LAU) in Beirut. His main research interests are in the areas of Artificial Intelligence, cybersecurity, Internet of Things and Federated Learning.
\end{IEEEbiography}

\vskip 0pt plus -1fil
\begin{IEEEbiography}[{\includegraphics[width=1in,height=1.25in,clip,keepaspectratio]{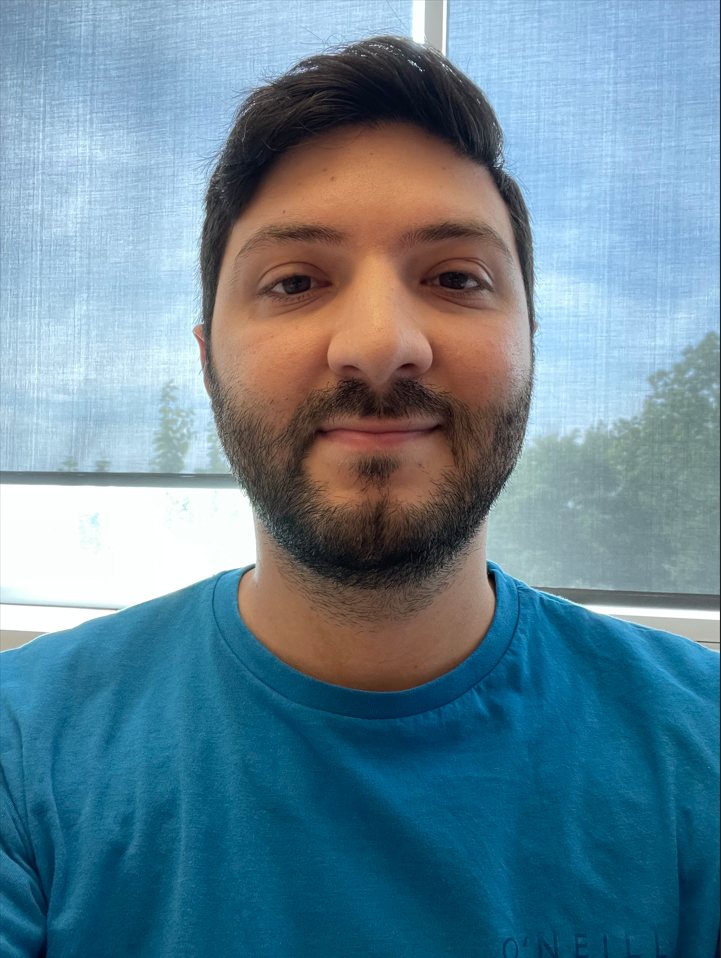}}]%
{\textbf{\textit{\textbf{Omar Abdel Wahab}}}}
received the M.Sc. degree in computer science from Lebanese American University, Beirut, Lebanon, in 2013, and the Ph.D. degree in information and systems engineering from Concordia University, Montreal, QC, Canada. He is Assistant Professor with the Department of Computer and Software Engineering, Polytechnique Montréal, Canada. From January 2019 till July 2022, he was Assistant Professor at Université du Québec en Outaouais, Gatineau, QC, Canada. In 2017, he did a Postdoctoral Fellowship with the École de Technologie Supérieure, Montreal, where he worked on an industrial research project in collaboration with Rogers and Ericsson. His current research activities are in the areas of cybersecurity, Internet of Things and artificial intelligence. Dr. Wahab is a recipient of many prestigious grants from prestigious agencies in Canada, such as the Natural Sciences and Engineering Research Council of Canada and Mitacs. He is a TPC member and a publicity chair of several prestigious conferences and a reviewer of several highly ranked journals.
\end{IEEEbiography}

\vskip 0pt plus -1fil

\begin{IEEEbiography}[{\includegraphics[width=1in,height=1.25in,clip,keepaspectratio]{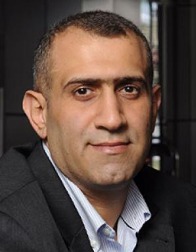}}]%
{\textbf{\textit{\textbf{Hadi Otrok}}}}
received his Ph.D. in ECE from Concordia University. He holds a Full Professor position in the department of Electrical Engineering and Computer Science (EECS) at Khalifa University. Also, he is an Affiliate Associate Professor in the Concordia Institute for Information Systems Engineering at Concordia University, Montreal, Canada, and an Affiliate Associate Professor in the Electrical department at Ecole de Technologie Superieure (ETS), Montreal, Canada. His research interests include the domain of blockchain, reinforcement learning, crowd sensing and sourcing, ad hoc networks, and cloud security. He co-chaired several committees at various IEEE conferences. He is also an Associate Editor at IEEE Transactions on Services Computing, IEEE Transactions on Network and Service Management (TNSM), Ad-hoc networks (Elsevier), and IEEE Network. He also served from 2015 to 2019 as an Associate Editor at IEEE Communications Letters.
\end{IEEEbiography}
\vskip 0pt plus -1fil

\begin{IEEEbiography}[{\includegraphics[width=1in,height=1.25in,clip,keepaspectratio]{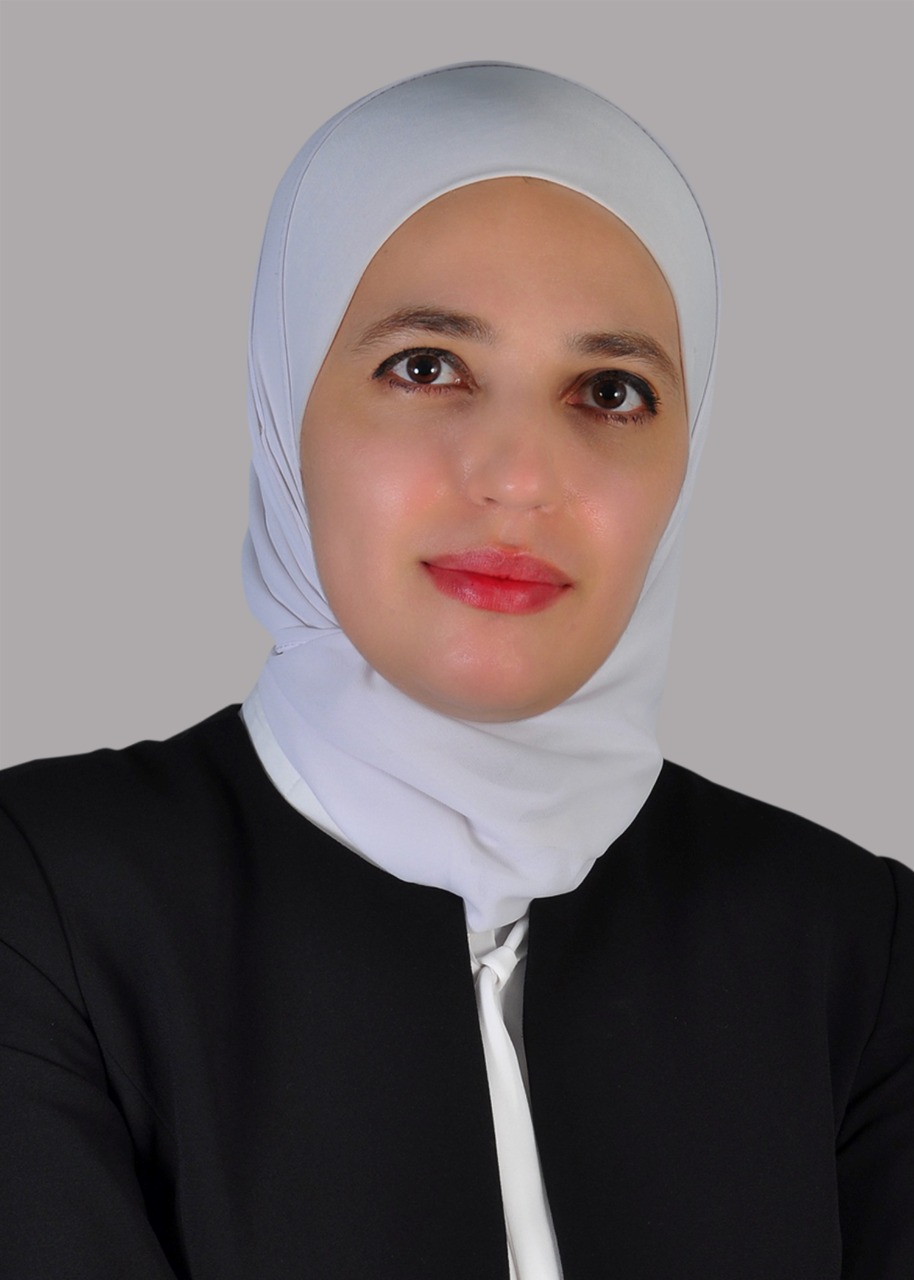}}]%
{\textbf{\textit{\textbf{Safa Otoum}}}}
(M’19) is an assistant professor of computer engineering in the College of Technological Innovation (CTI), Zayed University, United Arab Emirates. She received her M.A.Sc. and Ph.D. degrees in computer engineering from the University of Ottawa, Canada, in 2015 and 2019, respectively. Her research interests include blockchain applications, applications of ML and AI, IoT, and intrusion detection and prevention systems. She is a registered Professional Engineer (P.Eng.) in Ontario.
\end{IEEEbiography}
\vskip 0pt plus -1fil

\begin{IEEEbiography}[{\includegraphics[width=1in,height=1.25in,clip,keepaspectratio]{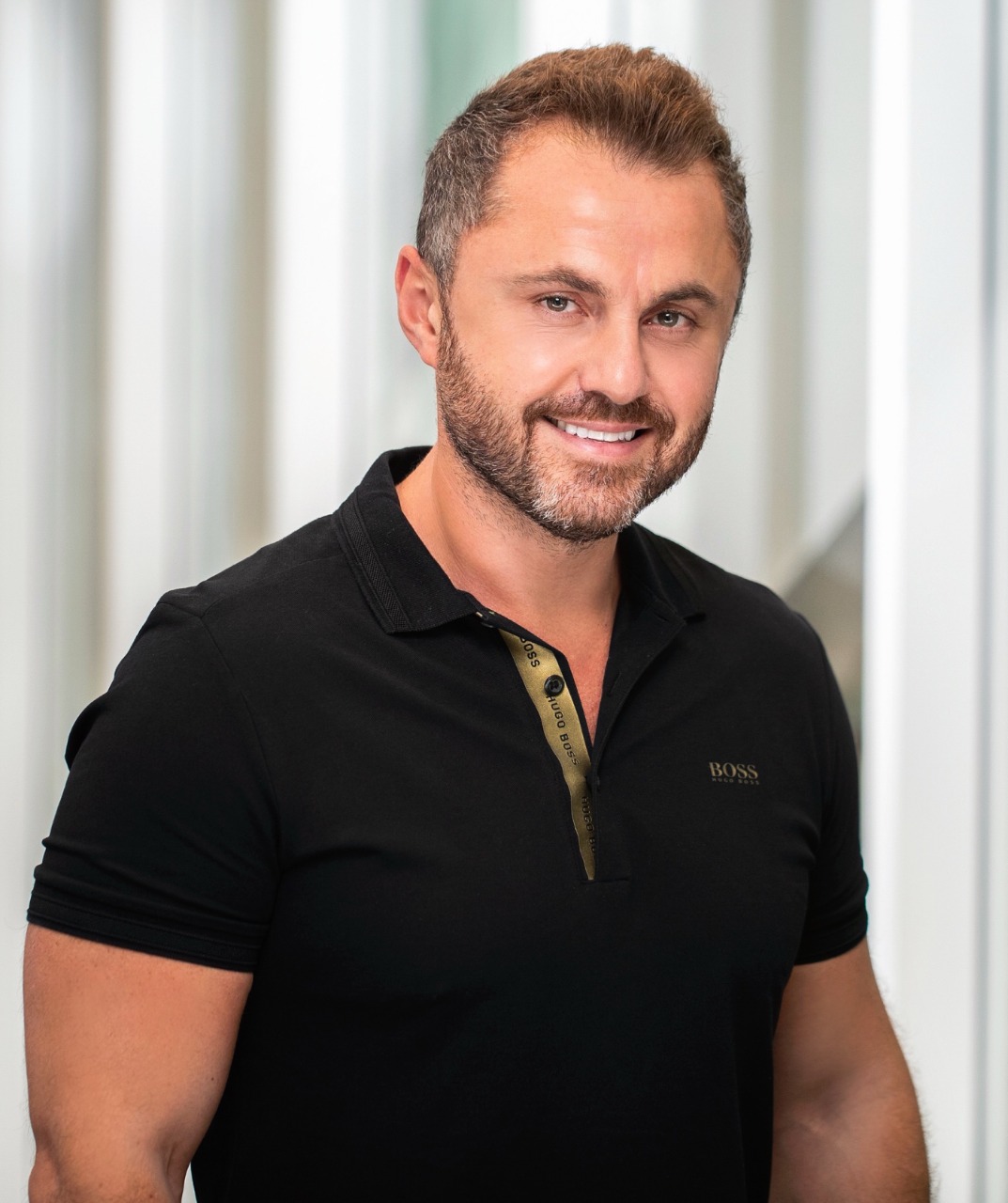}}]%
{\textbf{\textit{\textbf{Azzam Mourad}}}}received his M.Sc. in CS from Laval University, Canada (2003) and Ph.D. in ECE from Concordia University, Canada (2008). He is currently Professor of Computer Science and Founding Director of the Cyber Security Systems and Applied AI Research Center with the Lebanese American University, Visiting Professor of Computer Science with New York University Abu Dhabi and Affiliate Professor with the Software Engineering and IT Department, Ecole de Technologie Superieure (ETS), Montreal, Canada. His research interests include Cyber Security, Federated Machine Learning, Network and Service Optimization and Management targeting IoT and IoV, Cloud/Fog/Edge Computing, and Vehicular and Mobile Networks. He has served/serves as an associate editor for IEEE Transactions on Services Computing, IEEE Transactions on Network and Service Management, IEEE Network, IEEE Open Journal of the Communications Society, IET Quantum Communication, and IEEE Communications Letters, the General Chair of IWCMC2020, the General Co-Chair of WiMob2016, and the Track Chair, a TPC member, and a reviewer for several prestigious journals and conferences. He is an IEEE senior member.
\end{IEEEbiography}

\vskip 0pt plus -1fil
\begin{IEEEbiography}[{\includegraphics[width=1in,height=1.25in,clip,keepaspectratio]{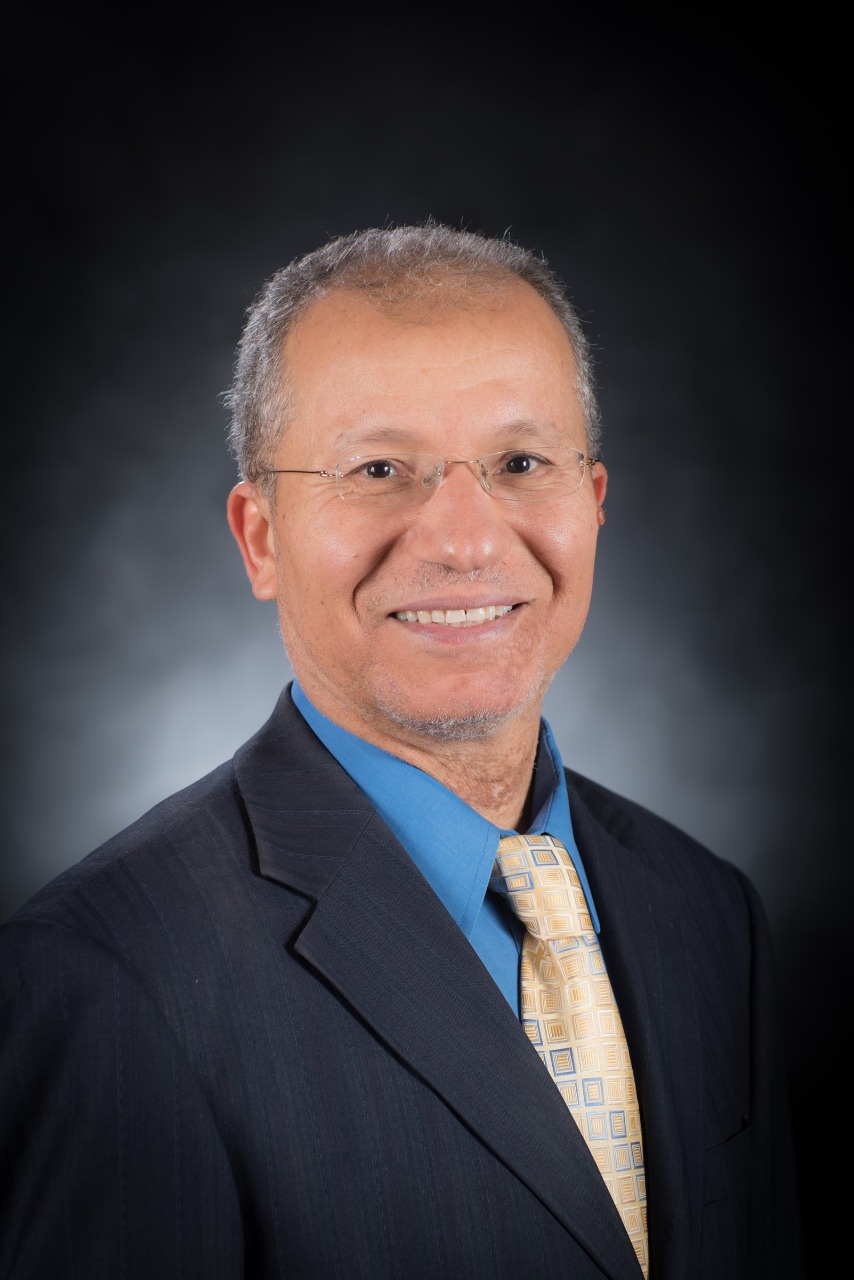}}]%
{\textbf{\textit{\textbf{Mohsen Guizani}}}}
(Fellow, IEEE) received the BS (with distinction), MS and PhD degrees in Electrical and Computer engineering from Syracuse University, Syracuse, NY, USA in 1985, 1987 and 1990, respectively. He is currently a Professor of Machine Learning and the Associate Provost at Mohamed Bin Zayed University of Artificial Intelligence (MBZUAI), Abu Dhabi, UAE. Previously, he worked in different institutions in the USA. His research interests include applied machine learning and artificial intelligence, Internet of Things (IoT), intelligent autonomous systems, smart city, and cybersecurity. He was elevated to the IEEE Fellow in 2009 and was listed as a Clarivate Analytics Highly Cited Researcher in Computer Science in 2019, 2020 and 2021. Dr. Guizani has won several research awards including the “2015 IEEE Communications Society Best Survey Paper Award”, the Best ComSoc Journal Paper Award in 2021 as well five Best Paper Awards from ICC and Globecom Conferences. He is the author of ten books and more than 800 publications. He is also the recipient of the 2017 IEEE Communications Society Wireless Technical Committee (WTC) Recognition Award, the 2018 AdHoc Technical Committee Recognition Award, and the 2019 IEEE Communications and Information Security Technical Recognition (CISTC) Award. He served as the Editor-in-Chief of IEEE Network and is currently serving on the Editorial Boards of many IEEE Transactions and Magazines. He was the Chair of the IEEE Communications Society Wireless Technical Committee and the Chair of the TAOS Technical Committee. He served as the IEEE Computer Society Distinguished Speaker and is currently the IEEE ComSoc Distinguished Lecturer. 
\end{IEEEbiography}

\end{document}